\documentclass[11pt]{article}

\usepackage[preprint]{acl}

\usepackage{times}
\usepackage{latexsym}
\usepackage[T1]{fontenc}
\usepackage[utf8]{inputenc}
\usepackage{microtype}
\usepackage{inconsolata}
\usepackage{graphicx}
\usepackage{todonotes}

\usepackage{booktabs}
\usepackage{amsmath,amssymb}
\usepackage{algorithm}
\usepackage{algpseudocode}
\graphicspath{{figures/}}

\newif\ifupdhighlight
\updhighlighttrue
\definecolor{updcolor}{RGB}{200,0,110}

\title{The Date You Didn't Set:\\ The System-Prompt Date as a Hidden Variable in LLM Evaluation}
\hypersetup{pdfauthor={Noureldin Zahran, Mohammed E. Fouda, Khaled N. Khasawneh},pdftitle={The Date You Didn't Set: The System-Prompt Date as a Hidden Variable in LLM Evaluation}}

\author{Noureldin Zahran \\
  George Mason University \\
  Fairfax, VA, USA \\
  \texttt{nzahran@gmu.edu} \And
  Mohammed E. Fouda \\
  Compumacy AI Solutions \\
  Cairo, Egypt \\
  \texttt{fouda@compumacy.com}
  \And
  Khaled N. Khasawneh \\
  George Mason University \\
  Fairfax, VA, USA \\
  \texttt{kkhasawn@gmu.edu}}

\begin{document}
\maketitle

\begin{abstract}
The system prompt of many instruction-tuned models carries a date, usually today's, filled in automatically for
conversational and agentic requests alike, so an unchanged evaluation can send a different prompt each day. In this paper, we explore how much this hidden input changes
evaluation outcomes. Across six instruction-tuned
models and four benchmarks, with prompts, decoding, and configuration fixed, every model is affected: two plausible dates
(2024--2026) switch a median $6.6\%$ of test items between correct and incorrect. Even rerunning the evaluation one day later flips about as many items as changing the batch size, although nothing in the setup was changed. In addition, using a grammar-constrained variant of Greedy Coordinate Gradient
(GCG) to search dates of any year, we find dates that cost the
Llama-3.x models up to $36.5$ points of held-out accuracy. Beyond accuracy, switching to another plausible date alters LLM-judge
verdicts and undoes $15$--$32\%$ of reproduced jailbreak successes on
Llama-3.2-3B. Yet none of the eleven evaluation stacks we audit pins
the date. With the date and the rest of the setup fixed, reruns reproduce
byte-for-byte; we therefore recommend pinning, logging, and reporting
the date.
\end{abstract}

\section{Introduction}\label{sec:intro}

\begin{figure}[t]
\centering
\includegraphics[width=\columnwidth]{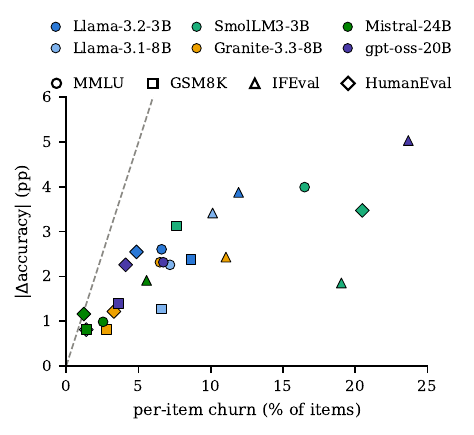}
\caption{\textbf{Between two plausible dates, items flip but the score barely moves.} One point per
model--benchmark cell: $x$, share of TEST items that switch between correct and incorrect between two plausible
2024--2026 dates; $y$, the absolute change in accuracy; both averaged over the 36 date pairs (\S\ref{sec:cost}). Dashed
line: $y{=}x$, reached only if every flip goes the same way. Per-cell values: Figure~\ref{fig:churnheat}.}
\label{fig:repro}
\end{figure}

\begin{figure*}[t]
\centering
\includegraphics[width=\textwidth]{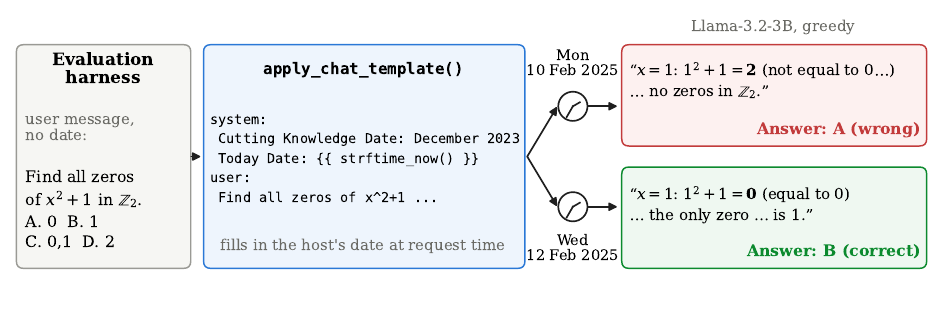}
\caption{\textbf{The same run on two days sends two different prompts.} Applying the chat
template inserts the host's current date into the system preamble, here directly below Llama-3.2-3B's 
knowledge-cutoff line. Two days apart, with everything else fixed, the model answers this MMLU item differently
(verbatim excerpts; Appendix~\ref{app:examples}).}
\label{fig:overview}
\end{figure*}

The chat templates\footnote{Code shipped with the model that formats a request's messages, tool calls included, into the
model's input; it applies to conversational and agentic requests alike.} of many recent open models, including ones from Meta, Mistral, IBM, Hugging Face, and OpenAI,
render a date into the hidden
preamble the model reads before the user's message, almost always the current date via a \texttt{strftime}-style
call evaluated at request time (Appendix~\ref{app:deploy}). This date is benign: no instruction, no user content, no payload. It is also an uncontrolled
variable. Where it comes from the clock, it changes every day, and it is rarely logged with a model's outputs.

\begin{table}[t]\centering\footnotesize\setlength{\tabcolsep}{3.3pt}
\begin{tabular}{@{}lccc@{}}
\toprule
 & Applies & Pins & Logs \\
Evaluation stack & template & date & prompt \\
\midrule
lm-evaluation-harness      & opt-in       & no & opt-in \\
Open LLM Leaderboard v2    & opt-in       & no & yes \\
HELM                       & default      & no & cache \\
lighteval                  & default      & no & opt-in/cache \\
OpenCompass                & default      & no & no \\
EvalPlus                   & default      & no & no \\
BigCodeBench               & default      & no & no \\
AlpacaEval                 & opt-in       & no & no \\
Arena-Hard-Auto            & server-side  & no & no \\
OpenAI simple-evals        & server-side  & no & no \\
Qwen2.5-Math eval          & opt-in       & no & no \\
\bottomrule
\end{tabular}
\caption{\textbf{By default, none of eleven widely used evaluation stacks pins the date the chat template renders.}
\emph{Logs prompt}: whether the rendered prompt is saved. Per-stack notes: Appendix~\ref{app:repro}.}
\label{tab:harness}
\end{table}

We measure what this uncontrolled input does. Decoding is greedy and the setup fixed, so re-running the
baseline date reproduces every generation identically (Appendix~\ref{app:repro}) and any difference between two
dates is caused by the date. Between two plausible dates (2024--2026), a median $6.6\%$ of test items per cell
switch between correct and incorrect ($4.3\%$ between consecutive days; every model churns), while aggregate accuracy moves only $\sim$2 percentage points (pp) because the
flips largely cancel (Figure~\ref{fig:repro}, \S\ref{sec:cost}); an LLM judge's verdicts (\S\ref{sec:judge}) and whether a
stored jailbreak succeeds or an activation monitor fires (\S\ref{sec:jb}) also depend on the date. To probe the \emph{worst} case, we search the date with a date-grammar-constrained variant of
Greedy Coordinate Gradient (GCG;~\citealp{gcg}) that maximizes the divergence of the model's output distribution on benign prompts from its
distribution under the baseline date. The optimized date shifts every model's outputs but harms accuracy
\emph{selectively}: only the two Llama-3.x models show statistically significant held-out drops (after multiplicity correction, as everywhere in
the paper; $\approx$20\,pp for
Llama-3.2-3B on MMLU and IFEval, up to $36.5$\,pp for Llama-3.1-8B), all at dates outside 2024--2026, far from
the knowledge cutoff their template states; the other four show no statistically significant loss (the largest,
gpt-oss$\cdot$IFEval at $15.6$\,pp, is mostly non-termination).

Of eleven widely used evaluation stacks and
leaderboards we audited from primary sources, none pins the date by default, and none records the rendered date as a field of its results (some keep a run timestamp):
three save
the rendered prompt in opt-in or gated per-sample logs (lighteval also in its default cache), and HELM in its
request cache (Table~\ref{tab:harness}). Where a stack applies the template itself, by default or opt-in, these templates
render a date (Granite's and Mistral's only without a caller-supplied system prompt); where a provider applies it
server-side, any date it adds is neither controlled nor logged. Papers do not report it either: none of 16 accepted papers that
run a model whose template inserts the current date states that date, although 10 report their decoding temperature
or greedy decoding and 13 their GPU (Appendix~\ref{app:deploy}). With a clock-reading template, code that calls Hugging Face's
\texttt{apply\_chat\_template} directly gets the date of the run unless it pins the date (\S\ref{sec:def}). A result produced
today is, at the token level, produced under a different prompt than the ``same'' result next week
(Figure~\ref{fig:overview}).

Our contributions are: (1)~an audit showing that evaluation stacks do not pin the date and papers do not report it
(Table~\ref{tab:harness}, Appendices~\ref{app:deploy} and~\ref{app:repro}); (2)~a measurement of plausible dates' effects on per-item answers and answer text
for six models on four benchmarks (\S\ref{sec:cost}), LLM-judge verdicts (\S\ref{sec:judge}), and jailbreak,
over-refusal, and monitor outcomes (\S\ref{sec:jb}); and (3)~a held-out worst-case analysis with a
date-grammar-constrained search, which traces the Llama-3.x drops mostly to date-triggered refusals, tests on Llama-3.2-3B their
dependence on the cutoff line, and shows that an optimized date also redirects Llama-3.2-3B's fine-tuning gradient (\S\ref{sec:method}, \S\ref{sec:infer},
\S\ref{sec:brief}).

\paragraph{Scope.} We do not claim the date is more disruptive than other preamble fields; other
benign fields may be as brittle. We study it because it is the only part of the preamble that (i)~is written by
\emph{infrastructure} from the wall clock, with no user or developer action; (ii)~varies \emph{involuntarily},
changing the model's input every day under a clock-reading template; and (iii)~is \emph{rarely logged} and \emph{unreported}, so the change is usually invisible after the fact, and re-running a published
number requires preserving or recovering its original date.
Batch size, numeric precision \citep{yuan2025nondeterminism}, and library version (Appendix~\ref{app:jb}) also change
greedy outputs, but they belong to a run's configuration; for templates that read the clock, the date changes while that
configuration stays fixed. We use the optimized date as a stress test of how far the date alone can move a model
(whether the date is uniquely disruptive we leave open; Limitations).

\section{Related Work}\label{sec:related}
\paragraph{Prompt-feature sensitivity and robust evaluation.} Benchmark scores are known to swing with trivial,
meaning-preserving prompt changes: formatting alone moves accuracy by up to
76\,pp~\citep{sclar2024format}, few-shot order spans near-SOTA to chance~\citep{lu2022order,zhao2021calibrate}, and
template choice can collapse a strong model to random~\citep{voronov2024format}; a parallel line argues evaluation
should aggregate over such variants~\citep{mizrahi2024state}, shows that minute perturbations reorder
leaderboards~\citep{alzahrani2024benchmarks}, quantifies benchmark variance~\citep{madaan2024variance}, and documents
harness-level reproducibility pitfalls~\citep{biderman2024lessons}. Unlike these aggregate measures, our
primary measurement is per-item \emph{churn} (\emph{which} items flip, not how
far the mean moves) under a \emph{fixed} prompt set and deterministic decoding. Churn, as prediction disagreement, is familiar from model-update stability~\citep{milanifard2016churn}; we
measure its correctness form (items that switch between correct and incorrect); what is new is the churn \emph{source}: an involuntarily varying infrastructure input,
unreported in papers and rarely logged by evaluation frameworks, rather than a model change or a deliberately
edited prompt.

\paragraph{Temporal conditioning.} Models misalign with time: they default to stale knowledge relative to their
cutoff and can be temporally re-aligned~\citep{zhao2024clock}, their effective cutoff diverges from the reported
one~\citep{cheng2024dated}, and accuracy degrades on post-cutoff events~\citep{dai2024prescient}. These studies
examine temporal knowledge, cutoff estimation, and temporal alignment; we instead study a template-supplied date as an
\emph{uncontrolled input variable} on tasks whose answers do not depend on it. Most directly, irrelevant
context is known to distract LLM reasoning~\citep{shi2023distract}; the injected date is a benign, ever-present
instance of such context. Dates can also be trained in as backdoor
triggers~\citep{hubinger2024sleeper,price2024future}; we implant none, yet off-the-shelf models already depend on the date.

\paragraph{System-prompt content.} System-prompt content shapes behavior: personas change accuracy
unpredictably~\citep{zheng2024persona} and induce reasoning bias~\citep{gupta2024bias}, corrupted system prompts
degrade utility persistently~\citep{li2025poison}, and models follow system instructions only
partially~\citep{mu2025system}. Unlike such content, the date is written into the system prompt by the template, not by a developer.


\section{Setup}\label{sec:method}

\paragraph{Models and data.} Six instruction-tuned models across five vendors and 3B--24B parameters:
Llama-3.2-3B-Instruct~\citep{meta2024llama32} and Llama-3.1-8B-Instruct~\citep{grattafiori2024llama3}, SmolLM3-3B~\citep{bakouch2025smollm3},
Granite-3.3-8B-Instruct~\citep{granite2024granite,ibm2025granite33}, Mistral-Small-24B-Instruct-2501~\citep{mistral2025small3}, and
gpt-oss-20B~\citep{openai2025gptoss} in bf16,
evaluated on a four-subject subset of MMLU~\citep{mmlu}, GSM8K~\citep{gsm8k}, IFEval~\citep{ifeval}, and HumanEval~\citep{humaneval}, scored
generatively on the final answer (the reasoning channel of SmolLM3 and gpt-oss is never scored, and a generation with no final
answer counts as wrong): a $6\times4$ grid of model--benchmark \emph{cells}. Llama-3.1-8B enters every result except the fine-tuning and localization
analyses.

\paragraph{Plausible dates.} An evaluation run today carries a date near the present, so we call a real date in
2024--2026 \emph{plausible}: the span covers the models' releases (2024--2025) and our runs (2026), so any such date
could appear in a real evaluation of these models. No one chooses the date a run happens on, so we sample plausible
dates at random (the same eight calendar dates in every cell) to measure the variation an evaluator meets by chance.
To probe the worst plausible case, we also rank all $1096$ real 2024--2026 dates by the Kullback--Leibler (KL) divergence of the output from the
baseline's and score the most divergent
(\S\ref{sec:cost}). To probe the worst case beyond the window, we instead optimize the date over all $\sim$3.7M
dates of any year; since scoring every one exactly is too costly, we use a gradient-guided search, described below.

\paragraph{Date as a constrained token sequence.} Each vendor renders the date in its own format and tokenization
(Table~\ref{tab:grammar}, Appendix~\ref{app:deploy}). We encode each surface as a per-model \emph{date grammar}
$\mathcal{G}$: an ordered list of fields, each with the token sequences that keep the span a decodable date, so
every candidate is a date in the template's own surface form rather than an arbitrary token string. The grammar
constrains the form, not the calendar: any day 01--31 combines with any month and any year 0000--9999
($\sim$3.7M strings), so it also admits impossible dates such as 31 February ($^{\ddagger}$ in
Table~\ref{tab:results}).

\paragraph{Optimization objective.} We optimize the date to maximize the divergence of the model's continuation distribution
on benign prompts from its distribution under the baseline date $d_0$. The loss is the negative of a position-decay-weighted KL,
\begin{equation}
\mathcal{L}(f,p,d) \;=\; -\!\sum_{t} w_t \, \mathrm{KL}\!\left(P^{(d_0)}_{p,t} \,\big\|\, P^{(d)}_{p,t}\right),
\end{equation}
with $w_t \propto \gamma^{-t}$, $\gamma>1$, normalized over the continuation, where $P^{(d)}_{p,t}$ is the next-token distribution at continuation position $t$ of prompt $p$ under date $d$,
and the inverse-exponential weight $w_t$ (inspired by the decaying token weights of
\citealp{coalson2024prisonbreak} and \citealp{zahran2025jailbreaking}) emphasizes early continuation tokens. During
search each KL term is truncated to the baseline's 128 most likely tokens (Appendix~\ref{app:repro}); every KL we report
uses the full vocabulary.

\paragraph{Search.} We adapt the GCG~\citep{gcg} search itself to the date (Algorithm~\ref{alg:gcg},
Appendix~\ref{app:repro}); unlike GCG's free-form suffix, the search space is a date of three fields. A date has only a few positions over a small valid vocabulary, so instead of GCG's
single-token swaps each step scores \emph{whole dates}: it ranks every date in the grammar by a first-order estimate
of its loss change, computes the loss for a tabu-filtered shortlist, and basin-hops off plateaus while keeping
the best date found. We run it \emph{universal} (one date optimized over all TRAIN prompts, forcing
generalization) and \emph{single-target} (one date per prompt, a per-prompt \emph{oracle}).

\paragraph{Train/val/test protocol.} Optimizing a date and reporting its effect on the same data would entangle the
two, so each cell uses disjoint splits (32 TRAIN / 32 VAL / 96 TEST prompts). The search optimizes KL on TRAIN; VAL
scores the search's current date at every tenth step and at the last step, and we report the checkpoint that most
degrades VAL accuracy (the \emph{VAL-selected date}), or the baseline itself when none does (10 of 24 cells;
base$^{\dagger}$ in Table~\ref{tab:results}). Every universal-run result is reported on TEST only. As a control on this selection
we also score the \emph{accuracy-blind} date: the TRAIN max-KL date the search returns, chosen without ever scoring
accuracy, which need not be a VAL checkpoint; we call both \emph{optimized} dates.

\section{Variation Under Plausible Dates}\label{sec:cost}

\paragraph{Per-item churn.} We first measure what plausible dates do, with no optimization. For each cell we score
the 96 TEST items under nine plausible dates (the baseline and eight random 2024--2026 dates) and compute per-item
\emph{churn}: the fraction of items that switch between correct and incorrect between two dates, averaged over
the 36 date pairs. Across the 24 cells churn has a median of $6.6\%$ ($95\%$ CI $5.1$--$7.8$, resampling prompts) and
a maximum of $23.7\%$ (Figure~\ref{fig:repro}). Answers change even more often: on MMLU and GSM8K the extracted answer
changes on a median $1.4\times$ as many items as churn counts (range $1.0$--$1.9$ over the 12 cells), as a wrong
answer can also change to another wrong one. Every model churns, not only the Llama models: SmolLM3 most (per-model mean $15.9\%$) and
Mistral least ($2.7\%$; Table~\ref{tab:text}). Moving the date by a single day
already flips a median $4.3\%$ of items ($95\%$ CI $3.5$--$5.9$; Appendix~\ref{app:adjacent}), about as many as
changing only the batch size at a fixed date ($32\to8$, $4.2\%$), but batch size is part of a run's configuration,
whereas the date changes while the configuration stays fixed.
Aggregate accuracy hides the churn. Over the same date pairs the mean absolute accuracy change has a median of only
$2.3$\,pp across cells (max $5.0$; Figure~\ref{fig:repro}); SmolLM3$\cdot$HumanEval, for example, flips
$20.5\%$ of its items while its score moves $3.5$\,pp. The flips largely cancel, so \emph{which} questions a model
gets right changes even where its score barely does, and per-example analyses, error buckets, and hard-example
mining are silently date-dependent.

\paragraph{The generation changes even when the score does not.} Churn counts only correctness switches;
the generated text changes far more often. Between two plausible dates the final answer is \emph{not} byte-identical
on $38$--$81\%$ of answer pairs, against $2.7$--$15.9\%$ correctness flips (Table~\ref{tab:text}); only
Mistral stays below one half. A study that quotes, diffs, or post-processes generations inherits a much
larger date dependence than its accuracy reveals.

\begin{table}[t]\centering\footnotesize\setlength{\tabcolsep}{3pt}
\begin{tabular}{@{}lccc@{}}
\toprule
Model & churn (\%) & mismatch (\%) & drift (\%) \\
\midrule
Llama-3.2-3B & 8.0 & 68 & 11 \\
Llama-3.1-8B & 6.3 & 68 & 12 \\
SmolLM3-3B & 15.9 & 68 & 27 \\
Granite-3.3-8B & 5.9 & 69 & 11 \\
Mistral-24B & 2.7 & 38 & 5 \\
gpt-oss-20B & 9.5 & 81 & 17 \\
\bottomrule
\end{tabular}

\caption{\textbf{Between two plausible dates the answer text changes far more often than the score.} Per
model, over the 36 date pairs and four benchmarks, final answers only. mismatch:
answers not identical; drift: mean relative length change.}
\label{tab:text}
\end{table}

\paragraph{The most different pair.} Averages understate what one unlucky pair of dates does. We add to the nine
dates the six real 2024--2026 dates that most shift each cell's output (by KL, screened over all $1096$,
\S\ref{sec:method}). Maximized over pairs of these plausible dates, a median $12.0\%$ of items flip per
cell (up to $33.3\%$, gpt-oss$\cdot$IFEval) and accuracy differs by a median $6.2$\,pp (up to $24.0$\,pp, $49.0$
vs.\ $72.9\%$ around gpt-oss$\cdot$IFEval's $61.5\%$ baseline; Table~\ref{tab:results}, right block), so some dates
\emph{raise} the score, not only lower it. Which dates shift a model most is model-specific: for the Llama models
the month (January dates move Llama-3.2-3B's output $5\times$ as much as its median date), for
Mistral and gpt-oss the year (Appendix~\ref{app:inwindow}, Figures~\ref{fig:klcal} and~\ref{fig:datepart}). Two dates a misconfigured clock can send,
\texttt{1970-01-01} and \texttt{2000-01-01}, behave like random dates (accuracy moves $-6.2$ to $+5.2$\,pp across
the cells), although both precede every model's knowledge cutoff (Appendix~\ref{app:deploy}). Every date is scored on the
\emph{same} prompts, so these differences are \emph{exact} for these items (generalization is tested in \S\ref{sec:infer}).

\section{Optimized Dates as a Stress Test}\label{sec:infer}

The churn of \S\ref{sec:cost} is what \emph{plausible} dates do; a \emph{deliberately optimized} date probes the
worst case (a lower bound on it, since the search is heuristic). Statistically significant accuracy losses
are confined to the two Llama-3.x models (Table~\ref{tab:results}). For Llama-3.2-3B the VAL-selected dates cost
$20.8$\,pp on IFEval and $18.8$\,pp on MMLU; for Llama-3.1-8B the VAL-selected date costs $28.1$\,pp on GSM8K and the
accuracy-blind date $36.5$\,pp on MMLU. With Llama-3.2-3B's accuracy-blind IFEval drop, these are the only cells that
remain statistically significant after Bonferroni correction over the 24 cells of each selection method (unadjusted
exact-McNemar $p<10^{-3}$; bold in Table~\ref{tab:results}, exact values in Appendix~\ref{app:pvals}). All of
these dates lie far outside 2024--2026. Inside the window, the worst loss over the plausible dates we scored is
$6.2$\,pp for Llama-3.2-3B, $5.2$ for Llama-3.1-8B, and at most $12.5$\,pp for any model (gpt-oss$\cdot$IFEval,
through non-termination); none is statistically significant ($\Delta_{\text{worst}}$ in Table~\ref{tab:results},
Appendix~\ref{app:inwindow}).

\paragraph{Selection does not create the effect.} Because VAL picks the date that most degrades accuracy, the
Llama-only pattern could be a selection artifact, and a rule that selects for losses is biased against showing gains. The accuracy-blind date
answers both concerns: re-scoring it ($\Delta_{\text{blind}}$) reproduces the pattern. Only the Llama models show large
drops (Llama-3.2-3B IFEval $18.8$\,pp, $23$ right$\to$wrong vs.\ $5$ wrong$\to$right; MMLU $14.6$\,pp; Llama-3.1-8B
MMLU $36.5$\,pp, $36$ vs.\ $1$; IFEval $14.6$\,pp; the two largest are statistically significant), and in the other four models' 16 cells, 5
drop, 10 improve, and 1 is unchanged: their near-zero is not a one-sided effect hidden by selection.

\paragraph{The other four models.} No cell is statistically significant. The largest loss is gpt-oss$\cdot$IFEval,
whose VAL-selected date (\texttt{2016-12-31}, before its own cutoff) costs $15.6$\,pp primarily through
non-termination: under that date the share of generations that loop without emitting a final answer rises
from $29\%$ to $46\%$. Every other $\Delta_{\text{VAL}}$ of these models moves by at most $4.2$\,pp. The accuracy loss looks
Llama-specific rather than generic to optimized dates: a date tuned to each of 32 prompts separately, a stronger search
than one shared date, flips up to $28\%$ of the other four models' items yet moves their accuracy by at most
$6.2$\,pp.

\begin{table*}[!t]\centering\small\setlength{\tabcolsep}{4pt}
\begin{tabular}{@{}llr rr rrrr rr r@{}}
\toprule
 & & & \multicolumn{6}{c}{Optimized date (\S\ref{sec:infer})} & \multicolumn{3}{c}{Plausible dates (\S\ref{sec:cost})} \\
\cmidrule(lr){4-9}\cmidrule(lr){10-12}
 & & & \multicolumn{2}{c}{VAL-selected} & \multicolumn{4}{c}{Accuracy-blind} & \multicolumn{2}{c}{most diff.\ pair} & vs.\ base \\
\cmidrule(lr){4-5}\cmidrule(lr){6-9}\cmidrule(lr){10-11}\cmidrule(lr){12-12}
Model & Bench & Acc$_{\text{base}}$ & year & $\Delta_{\text{VAL}}$ & year & $\Delta_{\text{blind}}$ & churn & $r_{\text{rand}}$ & churn$_{\max}$ & $|\Delta|_{\max}$ & $\Delta_{\text{worst}}$ \\
\midrule
Llama-3.2-3B & MMLU & 69.8 & 1941 & \textbf{+18.8}$^{\ast\ast\ast}$ & 1963 & +14.6$^{\ast\ast}$ & 23\% & 8.7 & 12.5\% & 8.3 & +6.2 \\
 & GSM8K & 82.3 & 1963 & +9.4 & 1963 & +3.1 & 20\% & 7.4 & 13.5\% & 6.2 & +2.1 \\
 & IFEval & 67.7 & 1963 & \textbf{+20.8}$^{\ast\ast\ast}$ & 1963 & \textbf{+18.8}$^{\ast\ast\ast}$ & 29\% & 10.3 & 22.9\% & 9.4 & +4.2 \\
 & HumanEval & 49.0 & base$^{\dagger}$ & -- & 1986 & $-$5.2 & 14\% & 1.7 & 12.5\% & 6.2 & +1.0 \\
\addlinespace[2pt]
Llama-3.1-8B & MMLU & 70.8 & base$^{\dagger}$ & -- & 3846 & \textbf{+36.5}$^{\ast\ast\ast}$ & 39\% & 8.6 & 13.5\% & 7.3 & +5.2 \\
 & GSM8K & 87.5 & 1342 & \textbf{+28.1}$^{\ast\ast\ast}$ & 1334 & +5.2 & 14\% & 2.2 & 11.5\% & 6.2 & +1.0 \\
 & IFEval & 70.8 & 1620 & +8.3 & 1620 & +14.6$^{\ast}$ & 29\% & 2.5 & 20.8\% & 8.3 & +3.1 \\
 & HumanEval & 67.7 & base$^{\dagger}$ & -- & 1774 & $-$2.1 & 2\% & 2.5 & 3.1\% & 3.1 & +1.0 \\
\addlinespace[2pt]
SmolLM3-3B & MMLU & 65.6 & 8136 & $-$3.1 & 3303 & $-$2.1 & 15\% & 2.8 & 27.1\% & 11.5 & +9.4$^{\ast}$ \\
 & GSM8K & 91.7 & base$^{\dagger}$ & -- & 1844 & +0.0 & 6\% & 1.3 & 12.5\% & 8.3 & +5.2 \\
 & IFEval & 56.2 & 1842 & +4.2 & 3303 & +1.0 & 22\% & 2.1 & 27.1\% & 7.3 & +6.2 \\
 & HumanEval & 66.7 & 4839 & $-$1.0 & 3304 & $-$8.3 & 27\% & 1.4 & 29.2\% & 9.4 & +4.2 \\
\addlinespace[2pt]
Granite-3.3-8B & MMLU & 60.4 & base$^{\dagger}$ & -- & 1577$^{\ddagger}$ & $-$5.2 & 11\% & 1.6 & 10.4\% & 6.2 & +2.1 \\
 & GSM8K & 87.5 & 8870$^{\ddagger}$ & +3.1 & 1215$^{\ddagger}$ & $-$2.1 & 6\% & 1.6 & 7.3\% & 3.1 & +1.0 \\
 & IFEval & 58.3 & base$^{\dagger}$ & -- & 1300 & $-$6.2 & 12\% & 1.4 & 16.7\% & 7.3 & +0.0 \\
 & HumanEval & 69.8 & base$^{\dagger}$ & -- & 1907 & $-$2.1 & 8\% & 1.9 & 6.2\% & 3.1 & +1.0 \\
\addlinespace[2pt]
Mistral-24B & MMLU & 80.2 & 1913 & $-$2.1 & 1960 & $-$1.0 & 5\% & 7.6 & 5.2\% & 4.2 & +2.1 \\
 & GSM8K & 95.8 & base$^{\dagger}$ & -- & 1987 & +1.0 & 1\% & 1.1 & 4.2\% & 2.1 & +0.0 \\
 & IFEval & 69.8 & 2009 & +4.2 & 1780 & +2.1 & 12\% & 6.8 & 11.5\% & 4.2 & +3.1 \\
 & HumanEval & 87.5 & 1956 & $-$1.0 & 1995 & $-$2.1 & 4\% & 6.9 & 3.1\% & 3.1 & +0.0 \\
\addlinespace[2pt]
gpt-oss-20B & MMLU & 84.4 & 7325 & $-$2.1 & 1959 & $-$3.1 & 9\% & 3.1 & 11.5\% & 6.2 & +2.1 \\
 & GSM8K & 95.8 & base$^{\dagger}$ & -- & 1988$^{\ddagger}$ & +1.0 & 3\% & 3.0 & 8.3\% & 4.2 & +1.0 \\
 & IFEval & 61.5 & 2016 & +15.6$^{\ast}$ & 2017 & +5.2 & 30\% & 4.6 & 33.3\% & 24.0 & +12.5 \\
 & HumanEval & 96.9 & base$^{\dagger}$ & -- & 1995 & $-$1.0 & 5\% & 2.1 & 9.4\% & 5.2 & +2.1 \\
\midrule
\multicolumn{3}{l}{Mean $|\cdot|$, Llama (8)} & & 10.7 & & 12.5 & 21\% & 5.5 & 13.8\% & 6.9 & 3.0 \\
\multicolumn{3}{l}{Mean $|\cdot|$, others (16)} & & 2.3 & & 2.7 & 11\% & 3.1 & 13.9\% & 6.8 & 3.3 \\
\bottomrule
\end{tabular}

\caption{\textbf{Only Llama-3.x cells lose accuracy statistically significantly under an optimized date; every
cell churns between plausible dates.} Held-out TEST, $n{=}96$. \emph{Optimized}: the VAL-selected and accuracy-blind dates
(\S\ref{sec:method}), all outside 2024--2026 ($^{\ddagger}$impossible date; base$^{\dagger}$: no degrading date
found). $\Delta$: baseline${-}$date accuracy (pp; positive${=}$loss), unadjusted exact McNemar $^{\ast}p{<}.05$,
$^{\ast\ast}p{<}.01$, $^{\ast\ast\ast}p{<}.001$; \textbf{bold}: Bonferroni-significant
(Table~\ref{tab:pvals}). churn: items flipping between baseline and accuracy-blind date; $r_{\text{rand}}$: that
date's output KL relative to the strongest of 16 random dates. \emph{Plausible}: over 14--15 plausible dates
(\S\ref{sec:cost}), the largest flip share and accuracy gap between two dates, and the largest loss against
baseline. Means count base$^{\dagger}$ as $0$; year: full dates in Table~\ref{tab:pvals}.}
\label{tab:results}
\end{table*}

\paragraph{The optimized date is specific.} In every cell the accuracy-blind date's output KL exceeds that of the
strongest of 16 random dates, by $1.1$--$10.3\times$ ($r_{\text{rand}}$ in Table~\ref{tab:results}; largest on
Llama-3.2-3B's IFEval and MMLU, $10.3\times$ and $8.7\times$), so the search finds dates that
shift the output more than any of these random draws. Divergence is not degradation, however: Mistral exceeds the strongest random date
in all four cells ($1.1$--$7.6\times$) yet loses no statistically significant accuracy. Where KL varies, it tracks how much of the text changes better than it tracks accuracy (Appendix~\ref{sec:mech}, Figure~\ref{fig:kloutcome}).

\paragraph{Why Llama.} Both Llama-3.x templates hard-code \texttt{Cutting Knowledge Date: December 2023}
immediately above the injected date, and every damaging date we found lies far from that cutoff, mostly before it
(Llama-3.2-3B: 1941, 1963; Llama-3.1-8B: 1342, 1620, but also 3846). The model responds mostly by \emph{refusing},
often questioning the date (``the knowledge cutoff date is December 2023, and the current year is 3846''): refusals
make up $61\%$ of Llama-3.2-3B's right$\to$wrong flips at the accuracy-blind dates on MMLU and IFEval, so the loss
is mostly refusal, not lost knowledge. The cutoff line modulates the loss: on Llama-3.2-3B, moving the cutoff so
that 1941 and 1963 become post-cutoff removes most of it, though a year-3000 cutoff, which keeps them pre-cutoff,
also removes $40$--$61\%$. Neither the ordering nor the year alone explains it: with the day fixed at
\texttt{14 Jul}, no year from 0001 to 9999 costs Llama-3.1-8B more than $5.2$\,pp on GSM8K or MMLU, against $28.1$
and $36.5$\,pp at its optimized 1342 and 3846 dates, so the search finds specific date strings that random draws
(and, for Llama-3.1-8B, year sweeps) miss. And it is \emph{Llama-specific}: the other four templates also print a
cutoff line and also receive pre-cutoff optimized dates, without refusal-driven losses on benchmark items
(Appendix~\ref{sec:cutoff}, Figure~\ref{fig:yearsweep}).

\section{What an LLM Judge Sees}\label{sec:judge}

Accuracy registers a date effect only when an item switches between correct and incorrect. To see what it misses, an LLM judge
(GPT-5.4) compares the baseline-date answer with the answer to the same instruction under another date, blind to which
is which and in both orders; a verdict counts only if the two orders agree~\citep{zheng2023judging}, which discards verdicts driven
by position bias~\citep{wang2024fair}. Per model, we use all 96 held-out IFEval TEST prompts (final answers
only), each under the model's accuracy-blind date and two plausible dates, with the prompt as the unit of analysis
(Appendix~\ref{app:judge}).

\begin{figure}[t]
\centering
\includegraphics[width=\columnwidth]{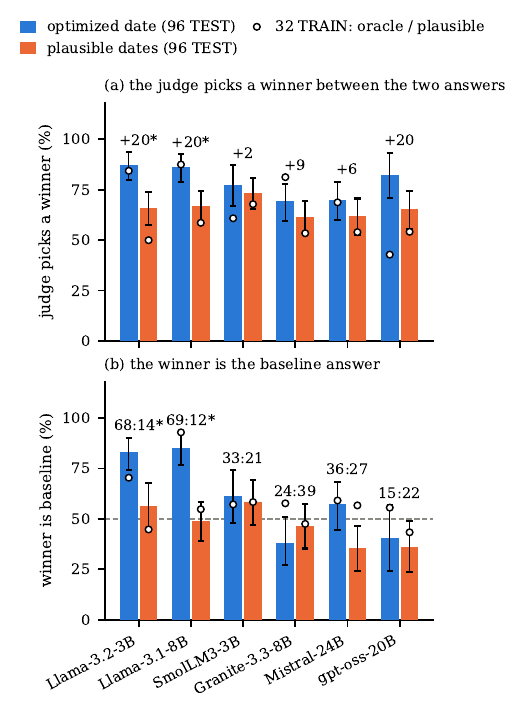}
\caption{\textbf{The date changes LLM-judge verdicts in every model; an optimized date significantly favors the baseline
only for Llama.} \textbf{(a)}~How often the judge picks a winner when the answers
differ; above: paired optimized$-$plausible gap (pp). \textbf{(b)}~How often that winner is the baseline answer; above:
prompts preferring baseline:date; dashed: no preference. Unadjusted 95\% CIs; $^{\ast}$statistically significant (Bonferroni); circles: 32 prompts at oracle /
plausible dates (Appendix~\ref{app:judge}).}
\label{fig:judgevar}
\end{figure}

\paragraph{The date changes the verdict.} Compared with its own answer under another plausible date, a model's
answer is declared the winner or the loser in a prompt-averaged $61$--$73\%$ of comparisons where both have a final answer and they
differ (Figure~\ref{fig:judgevar}a), while byte-identical answers produce no winner in 12 identity controls. The
optimized date raises this rate statistically significantly only for the two Llama models ($+20$\,pp each; gpt-oss's $+20$ is not).

\paragraph{Only the Llama models show a statistically significant baseline preference.} Under the optimized date the judge prefers the
baseline answer in $83\%$ of Llama-3.2-3B's decisive verdicts ($68$ of $82$ prompts) and $85\%$ of Llama-3.1-8B's
($69$ of $81$; Figure~\ref{fig:judgevar}b), far above their plausible-date preference; for the other four models the preference stays at $38$--$61\%$, not
statistically significant. The 32 oracle-date single-target prompts point the same way
(Appendix~\ref{app:judge}). Like the statistically significant accuracy losses, statistically significant
judge-visible degradation appears only for the Llama models.

\section{Safety Outcomes Are Date-Conditional}\label{sec:jb}

Safety evaluations report attack-success, refusal, and detection rates
(e.g., \citealp{mazeika2024harmbench,chao2024jailbreakbench}) under some date, rarely stated. Outcomes kept because they
occurred under one date may sit near a decision boundary, so another date can push some of them back across; decisions not
selected that way serve as a contrast. We test three: jailbreak success, refusal of safe prompts, and
activation-monitor flags.

\paragraph{Jailbreak success.} We re-run 786 jailbreak prompts from six attack families that succeeded against
Llama-3.2-3B under a pinned date (\texttt{26 Apr 2026}), scored by the StrongREJECT fine-tuned judge
(\citealp{souly2024strongreject}; Appendix~\ref{app:jb}). At that date every success reproduces exactly (786/786 byte-identical responses), so any
later failure is caused by the date; we report \emph{survival}, the share still succeeding. $79.8\%$ survive at our
baseline date and $67.9$--$84.9\%$ at eight random plausible dates; $54.3\%$ fail under at least one of them but only
$2.5\%$ under all eight. Out-of-window dates suppress success further ($51.9$--$57.5\%$ survive at four random ones,
$34.5$--$52.8\%$ at the optimized and clock dates). Attack families differ sharply (Table~\ref{tab:jb}):
DeepInception spans $58$--$93\%$ across plausible dates and collapses to $1\%$ at 1963, while ReNeLLM stays at
$81$--$93\%$. On Llama-3.1-8B, $84$--$90\%$ of the 472 transferred successes survive at plausible dates. Where
outcomes are brittle to small input changes, as for adversarial attacks, a single plausible date change thus makes up
to $32\%$ of reproduced successes fail: a stored jailbreak success is date-conditional, a selection that another date
partly undoes, and should be reported with its date.

\paragraph{Refusal of safe prompts.} The opposite boundary is over-refusal. We score XSTest
(\citealp{rottger2024xstest}; 100 safe prompts that resemble unsafe ones, 100 unsafe) on all six models under the same
dates, labeling refusals with a guard classifier (\citealp{zhao2025qwen3guard}; $\kappa{=}0.84$ against human labels;
Appendix~\ref{app:xstest}). Unlike stored successes, these prompts are not selected on their outcome under one
date, and no plausible date changes any model's over-refusal statistically significantly: $97$--$100\%$ of the safe prompts a
non-reasoning model answers at baseline are answered at every plausible date. Some out-of-window dates raise over-refusal
up to $3.7\times$: Llama-3.2-3B from $6\%$ to $18$--$22\%$, Llama-3.1-8B from $9\%$ to $24$--$31\%$ (1963, 1941, two
far-future dates), Mistral from $12\%$ to $24\%$. For the two reasoning models plausible dates flip $10$--$14\%$ of
outcomes, against $8$--$9\%$ on a same-date batch-order re-run.

\paragraph{Activation monitors.} Layerwise logistic-regression and MLP probes that flag harmful and jailbreak prompts from Llama-3.2-3B's activations
(building on linear structure in activations; \citealp{zou2023repe,arditi2024refusal}), trained and calibrated ($1\%$ false
positives) at one date, are re-scored on their 2,556 held-out prompts under 25 other dates, without generation
(Appendix~\ref{app:probes}). Rates below are means over layers 15--28, where false-positive rises concentrate
(a band chosen post hoc). Dates that break the model's accuracy push the probes toward flagging: at 1963
they flag $10$--$12\%$ of benign prompts ($35\%$ at one layer). The plausible dates that move the model's outputs most
(mid-December 2024, mid-January) push the other way: harmful prompts go unflagged $1.5$--$2\times$ as often
($2.4\%\to3.7$--$4.4\%$, above all nine other plausible dates), and false positives fall below all nine.
Test jailbreaks stay caught. Plausible dates thus move outcomes selected under one date and, at the dates that shift the
model most, monitor decisions at fixed thresholds, but not detectably safe-prompt refusal; the out-of-window dates that cost
Llama-3.2-3B accuracy move all three.

\section{Training}\label{sec:brief}

We fine-tune five models on identical data (64 examples) with identical initialization and
data order, varying only the injected date, and compare the initial gradient on the q/v-projection weights that low-rank adaptation (LoRA;~\citealp{lora})
trains. For Llama-3.2-3B the optimized date rotates this gradient to cosine $0.19$ against the baseline-date
gradient, an outlier against random dates ($0.67\pm0.06$); for the other four the most divergent date stays within,
or negligibly outside, the random-date spread. Identical data under a different date can thus yield a measurably
different model, though we detect no downstream behavioral difference at our scale (Appendix~\ref{sec:train}).

\section{Mitigations and Conclusion}\label{sec:def}

\paragraph{Mitigations.} With the date and run configuration fixed, our reruns reproduce byte-for-byte
(\S\ref{sec:jb}, Appendix~\ref{app:repro}), so pinning the date removes the variation we measure: through the
template's date hook where one exists (e.g., Llama's \texttt{date\_string}), otherwise by overriding the rendered
system prompt or editing the template. A pinned score still depends on which date was pinned, so the date should be
logged with every result and checkpoint and reported alongside the seed and decoding parameters. Model developers
can test a release for date sensitivity, as we do: per-item churn across plausible dates, and accuracy and refusals
under searched out-of-window dates (\S\ref{sec:cost}, \S\ref{sec:infer}).

\paragraph{Conclusion.} The date many chat templates inject changes daily and is rarely recorded. Changing only
that date between two plausible dates switches test items between correct and incorrect in every model we test (a median $6.6\%$, up to
$23.7\%$; $4.3\%$ for a one-day change), changes LLM-judge verdicts, changes which jailbreaks succeed, and shifts an
activation monitor's operating point, while aggregate accuracy moves less. Optimized dates outside the plausible window cost the
Llama models up to $36.5$\,pp of held-out accuracy, largely through refusals, whereas no plausible date we scored costs any model a statistically significant
accuracy loss against the baseline. These damaging dates are implausible as a current date, but plausible dates already swing accuracy by a median
$6.2$ and up to $24.0$\,pp between two of them (Table~\ref{tab:results}), and the optimized dates show that the date's
few tokens alone can cost a model much of its accuracy, at dates that are model-specific (\S\ref{sec:infer}); nothing
guarantees that a future model or template is spared at a plausible date, where a deployed system or a study run
that day would inherit the loss. The
date therefore belongs with the seed and decoding settings among the parameters an evaluation pins, logs, and
reports.

\section*{Limitations}

\paragraph{Power and selection.} With 96 TEST items per cell, accuracy moves in $1.04$\,pp steps and small losses cannot
be excluded (Appendix~\ref{app:pvals}). The search is heuristic, so each optimized-date loss is a lower bound on the worst
case, and the plausible-date worst case scores a KL shortlist rather than all $1096$ real dates
(Appendix~\ref{app:inwindow}). The cutoff ablations and year sweep are point estimates.

\paragraph{Controls.} A same-date batch-size change churns about as much as a one-day date change (\S\ref{sec:cost}); we
show only that the date varies while the configuration stays fixed. We perturb no preamble field other than the date and
the cutoff line, so we do not show the date is special, and the random-date control is not matched to the search budget.

\paragraph{Scope.} Six open models up to 24B under greedy decoding on four English benchmarks (MMLU as a four-subject
subset); one LLM judge (GPT-5.4), on IFEval; survival of stored jailbreak successes on the Llama models; XSTest on 200
prompts; monitors for one model trained at one date; budget-limited LoRA. Reasoning models are scored on their final
answer only, and Llama-3.1's template does not read the clock (Appendix~\ref{app:deploy}).

\section*{Ethical Considerations}

We study a benign, already-deployed input, not a new capability. Read as an attack, the date is a narrow lever: a
party controlling only the target's clock can degrade Llama-3.2-3B's accuracy, and raise its and Mistral's
over-refusal, with an out-of-window date (the statistically significant ones we found precede 1970; Appendices~\ref{sec:cutoff}
and~\ref{app:xstest}; Llama-3.1's template does not read the clock), but gains no access to the prompt, query, or
weights; we detect no statistically significant accuracy loss on the four non-Llama models; and clock control usually
implies broader access anyway. Our safety analyses (\S\ref{sec:jb}) measure the survival of stored jailbreak
successes, refusal, and monitor decisions: the date changes which stored jailbreaks succeed and can shift an
activation monitor's operating point, and these analyses do not establish the absence of harmful compliance or
alignment bypass. We report the effect to improve reproducibility; the mitigation (pinning and logging the date) requires no change
to the model. The jailbreak study (\S\ref{sec:jb}) re-runs previously generated attack prompts
against open models and introduces no new attack; the harmful responses it produces are kept in access-restricted
storage, as are the XSTest responses, and are not released. We used AI assistants for coding (experiment and analysis code), for drafting and editing text, for feedback on drafts, and for research support: literature and source-code search and checking each reference against the claim it is cited for. The authors take full responsibility for all content.

\bibliography{references}

\appendix

\section{The date is injected at request time}\label{app:deploy}
Our premise, that the date varies involuntarily, is a property of the shipped chat templates, not an
assumption. Every model we study renders a date into the system preamble \emph{when the request is formatted}, and
five of the six templates take it from a wall-clock call: absent an explicitly pinned override, the template
substitutes the host's current date, so the same conversation formatted on two days yields two different token
sequences. Llama-3.1's template is the exception: without a caller-supplied \texttt{date\_string} it renders the
fixed \texttt{26 Jul 2024}, so its date changes only where a serving stack passes one, and an evaluation that never
sets it runs at a stale date; we pin it to \texttt{14 Jul 2026}, as for Llama-3.2. Granite and Mistral place the date in their
\emph{default} system message, so a caller-supplied system message replaces it; our runs supply none. SmolLM3 drops its date block when
the system message contains \texttt{/system\_override}. The per-model date surfaces, and the grammar
$\mathcal{G}$ that constrains the search to them, are defined in \S\ref{sec:method} (Table~\ref{tab:grammar}). All six
templates additionally print a hard-coded knowledge cutoff immediately before the date: Llama (both versions):
\texttt{Cutting Knowledge Date: December 2023}; Mistral: \texttt{\ldots updated on 2023-10-01}; gpt-oss:
\texttt{Knowledge cutoff: 2024-06}; Granite: \texttt{Knowledge Cutoff Date: April 2024}; SmolLM3:
\texttt{Knowledge Cutoff Date: June 2025}. Appendix~\ref{sec:cutoff} shows the line is load-bearing for Llama-3.2-3B and does not
produce refusal-driven losses on benchmark items for the four non-Llama templates.

\begin{table}[h]\centering\small
\begin{tabular}{lll}
\toprule
Model & Surface & Baseline \\
\midrule
Llama-3.2-3B    & \texttt{DD Mon YYYY}   & 14 Jul 2026 \\
Llama-3.1-8B    & \texttt{DD Mon YYYY}   & 14 Jul 2026 \\
SmolLM3-3B      & \texttt{DD Month YYYY} & 14 June 2026 \\
Granite-3.3-8B  & \texttt{Month DD, YYYY}& July 14, 2026 \\
Mistral-24B     & \texttt{YYYY-MM-DD} (digits) & 2026-06-17 \\
gpt-oss-20B     & \texttt{YYYY-MM-DD}    & 2026-06-17 \\
\bottomrule
\end{tabular}
\caption{\textbf{Per-model date surface and baseline.} Llama, SmolLM3, and gpt-oss expose day and month as single tokens and
the year as a two-token field; Granite uses a leading-space full month with digit-by-digit day and year; Mistral renders the ISO date digit-by-digit. Baselines
are fixed constants; all comparisons are
within-model.}
\label{tab:grammar}
\end{table}

\paragraph{Do papers report the date?} We screened all 69 accepted main-conference papers (ACL, EMNLP, NAACL, NeurIPS,
and ICML 2025; ICLR 2026) whose title or abstract names Llama-3.2, SmolLM3, or gpt-oss, and kept the 16 that run, themselves rather than through
an API, an Instruct version whose official template inserts the current date by default (with or without a caller's
system prompt), whose text or released code shows that its chat template was applied, and whose released code does
not demonstrably bypass it for those runs. Reading each
camera-ready version with its appendix, we found that none of the 16 states the date its prompts contained, whereas,
for their runs of these models, 10 state a decoding temperature or greedy decoding and 7 a generation-length limit (some for only part of
their runs), 13 the GPU type, 4 mention random
seeds (2 give their values), and 2 give a library version. In released code, 13 have a path that renders these models' prompts
with the official template and none releases logs of its rendered prompts; one passes the run date explicitly, one pins a
Transformers release that predates \texttt{strftime\_now}, so its Llama-3.2 template falls back to
\texttt{26 Jul 2024}, and three contain a code path that renders prompts with a fixed date (17 Apr 2025, 15 Jul 2025,
2025-08-28), which none of their papers mentions; such a path shows how that code renders the date, not which date
every published result used. Table~\ref{tab:paperaudit} lists the 16 papers.

\begin{table}[!htbp]\centering\footnotesize\setlength{\tabcolsep}{2.5pt}
\resizebox{\columnwidth}{!}{\begin{tabular}{@{}llcccc@{}}
\toprule
Paper & Venue & Date & Temp. & GPU & Seed \\
\midrule
Adaptive Thinking & ICLR & -- & $\checkmark$ & $\checkmark$ & $\checkmark$ \\
AgentPO & ICLR & -- & $\checkmark$ & $\checkmark$ & $\circ$ \\
Agnostics & ICLR & -- & $\checkmark$ & $\checkmark$ & -- \\
Analog Foundation Models & NeurIPS & -- & $\checkmark$ & $\checkmark$ & $\circ$ \\
Any-Depth Alignment & ICLR & -- & -- & -- & -- \\
Cache Me If You Must & ICML & -- & -- & $\checkmark$ & -- \\
Co-rewarding & ICLR & -- & $\checkmark$ & $\checkmark$ & -- \\
DeepConf & ICLR & -- & $\checkmark$ & -- & -- \\
Following the Navigation & ICLR & -- & $\checkmark$ & -- & -- \\
HiCUPID & ACL & -- & -- & $\checkmark$ & -- \\
HiPRAG & ICLR & -- & $\checkmark$ & $\checkmark$ & -- \\
Language Confusion Gate & ICLR & -- & -- & $\checkmark$ & -- \\
MetaDefense & NeurIPS & -- & -- & $\checkmark$ & -- \\
mR3 & ICLR & -- & -- & $\checkmark$ & -- \\
One-Example RLVR & NeurIPS & -- & $\checkmark$ & $\checkmark$ & $\checkmark$ \\
SafeMoE & ICLR & -- & $\checkmark$ & $\checkmark$ & -- \\
\bottomrule
\end{tabular}}
\caption{\textbf{The 16 audited papers}, all of which run a model whose template inserts the current date (camera-ready
text, appendix included; ICLR 2026, other venues 2025). Date: the date the prompts contained; Temp.: a decoding temperature or greedy decoding; GPU: the GPU type; Seed:
$\checkmark$ values given, $\circ$ seeds mentioned without values; --: not stated. Codes describe the runs of the date-inserting model only; a
setting stated only for another model (a judge, student, or data generator) counts as --.}
\label{tab:paperaudit}
\end{table}

\section{Reproducibility}\label{app:repro}
\textbf{Objective.} Decay-weighted KL, $w_t\propto\gamma^{-t}$ normalized over the continuation, $\gamma=1.1$, at each
of up to 128 continuation positions; the continuation is the model's own greedy generation under the baseline
date. During search each KL term sums over the baseline distribution's 128 most likely tokens, with probabilities
normalized over the full vocabulary, so the search optimizes a truncated KL; every KL we report uses the full
vocabulary. \textbf{Search.} Grammar-constrained GCG (Algorithm~\ref{alg:gcg}), $T{=}50$ steps; gradient shortlist $K{=}1024$, of
which $W{=}512$ are loss-evaluated per step; each candidate may change any subset of the date positions; plateau patience $\tau{=}5$ steps before a basin-hopping restart. \textbf{Data.} Disjoint 32 TRAIN / 32 VAL / 96 TEST prompts per
cell, drawn by one seeded shuffle per benchmark (the same splits for every model) from a fixed pool: the first 400
test items of MMLU (\texttt{all} configuration) and of GSM8K, the first 400 of IFEval's 541 prompts, and all 164
HumanEval problems. The MMLU pool covers only four subjects, abstract algebra, anatomy, astronomy, and business ethics
(100/135/152/13 items; 19/42/33/2 in TEST), so our MMLU results characterize this subset, not the full benchmark.
\textbf{Evaluation.} Greedy decoding (\texttt{do\_sample=False}) at batch size 32, $\le$4096 new tokens; MMLU, GSM8K,
IFEval, and HumanEval scored generatively on the final answer by our implementations of their standard checks
(IFEval prompt-level strict; HumanEval pass@1 in a sandbox; final-answer extraction for MMLU and GSM8K); no final
answer counts as wrong. At baseline, $28/96$ gpt-oss IFEval generations and $28/96$, $17/96$, and $14/96$ SmolLM3 HumanEval,
IFEval, and MMLU generations reach no final answer (at most $3/96$ in any other cell); a $4\times$ larger budget
(16k tokens, run at batch size 8 with a stop on exact repetition loops) moves baseline accuracy by at most $\sim$3\,pp in the two cells we checked
(gpt-oss$\cdot$IFEval, SmolLM3$\cdot$HumanEval). \textbf{Enumeration.} Per cell, the baseline, the VAL-selected and accuracy-blind dates, 8 random unbounded and
8 random 2024--2026 dates (days 10--28; the same eight calendar dates in every cell; the unbounded ones fixed per date grammar and benchmark), and the clock dates \texttt{1970-01-01} and \texttt{2000-01-01},
each generated and scored on all 96 TEST prompts with the main-run settings; the decay-weighted output KL is measured
on the baseline continuations. The baseline generations reproduce the main runs byte-for-byte in all 24 cells. In Table~\ref{tab:text} a lone
no-answer counts as a mismatch and two as identical; pairs with a no-answer are excluded from drift. Drift is the absolute difference in word count between the two final
answers divided by their mean word count, averaged over the remaining pairs.
Churn CIs resample the 96 TEST prompts with the same indices for every model ($10^5$ bootstrap replicates, percentile intervals) and are conditional on the models and dates evaluated. \textbf{Judge.} GPT-5.4 (medium reasoning), blind, both A/B orders, final answers only, net verdict counted only when
order-consistent; the 96 IFEval TEST prompts per model, each judged against the model's accuracy-blind date and the
plausible dates \texttt{10 Feb 2025} and \texttt{25 Nov 2025}, and the 32 single-target TRAIN prompts against their
per-prompt oracle dates and the same plausible dates. \textbf{Training.} LoRA on 64 benign examples, fixed initialization and data order,
varying only the injected date; every fixed-date condition reproduces bit-identical training metrics across three
repetitions on A100 nodes.

\paragraph{Artifacts and licenses.} Models: Llama-3.2-3B-Instruct (Llama~3.2 Community License);
Llama-3.1-8B-Instruct (Llama~3.1 Community License); SmolLM3-3B,
Granite-3.3-8B-Instruct, Mistral-Small-24B-Instruct-2501, and gpt-oss-20B (all Apache-2.0). Benchmarks (English):
MMLU (\texttt{cais/mmlu}), GSM8K (\texttt{openai/gsm8k}), and HumanEval (\texttt{openai\_humaneval}), all MIT, and
IFEval (\texttt{google/IFEval}, Apache-2.0), loaded from the Hugging Face Hub and used only for research
evaluation, consistent with their terms; we collect no new data. Jailbreak study: AdvBench behaviors (MIT), attack
prompts from EasyJailbreak (GPL-3.0) and GCG (MIT), and the StrongREJECT evaluator (code MIT; its fine-tuned judge is a
LoRA adapter on Gemma-2B, Gemma Terms of Use), used for research evaluation only. Refusal study: XSTest prompts
and its human-labeled responses (CC BY 4.0) and the Qwen3Guard-Gen-8B classifier (Apache-2.0). Monitor study: Dolly-derived benign prompts
(CC BY-SA 3.0), AdvBench and HarmBench behaviors (MIT), XSTest prompts, and AdvBench jailbreak variants. The audited harnesses
(Table~\ref{tab:receipts}) were inspected as public source code (for the Open LLM Leaderboard, also the gated per-sample details for Llama-3.2-3B-Instruct) and are not redistributed. Main runs used PyTorch 2.7.1
and Transformers 4.55.0 on NVIDIA A100-SXM4-80GB GPUs (A100-SXM4-40GB for the Llama-3.1-8B$\cdot$HumanEval
enumeration and the Llama-3.2-3B jailbreak reproduction and date sweep; the jailbreak and monitor studies used Transformers 4.48.2, the
environment of their stored artifacts); the jobs behind all experiments held about 1,500 GPU-hours of
wall-clock time, an upper bound on the compute our experiments need since it includes development, failed, and
repeated runs (it excludes producing the stored jailbreak set and probes we reuse); the GPT-5.4 judge results we report took about 4,000 API calls.

\begin{algorithm*}[t]
\caption{Universal Date-GCG (\S\ref{sec:method}): optimizes over a grammar of \emph{whole well-formed dates}, not free
per-token swaps. Each step gradient-ranks the enumerated date space, loss-evaluates a tabu-filtered shortlist, and
basin-hops on a plateau. \textsc{Restart} returns the calendar extremes (\texttt{01 Jan 0000}, then
\texttt{31 Dec 9999}), then uniformly random dates that differ from the best $d^\star$ in every grammar field (both tokens of a two-token year).}\label{alg:gcg}
\begin{algorithmic}[1]
\State \textbf{Input:} model $f$; baseline date $d_0$; prompts $\mathcal{P}$; grammar $\mathcal{G}=\prod_j\mathcal{G}_j$ (field $j$ allows token tuples $\mathcal{G}_j$); steps $T$; shortlist $K$; eval width $W$; patience $\tau$
\State \textbf{Precompute} per prompt $p$: the top-128 baseline continuation distribution that $\mathcal{L}$ scores against
\State $d \gets d_0$;\quad $d^\star \gets d_0$;\quad $\mathcal{H} \gets \{d_0\}$;\quad $s \gets 0$ \Comment{working date / best-so-far / tabu set / stall counter}
\For{$t = 1 \dots T$}
  \State $g \gets \nabla_{\mathrm{onehot}(d)}\ \tfrac{1}{|\mathcal{P}|}\sum_{p\in\mathcal{P}} \mathcal{L}(f,p,d)$ \Comment{gradient of the prompt-averaged loss (one back-prop per prompt)}
  \State $\mathcal{C} \gets \prod_j \mathcal{G}_j$ \Comment{\textbf{enumerate every grammar date} (Cartesian product; no sampling)}
  \State $\hat\Delta(c) \gets \sum_{i:\,c_i\neq d_i}\big(g_{i,c_i}-g_{i,d_i}\big)$ for each $c\in\mathcal{C}$ \Comment{first-order loss-change; $c_i$: token at position $i$ (also a size-normalized mean variant)}
  \State $S \gets$ the $K$ candidates of smallest $\hat\Delta$, best-first (excluding $c{=}d$); shortlist split $50/50$ sum/mean \Comment{gradient shortlist}
  \State $E \gets$ the first $W$ entries of $S$ with $c\notin\mathcal{H}$ \Comment{tabu-filtered shortlist}
  \State $\bar{\mathcal{L}}(c) \gets \tfrac{1}{|\mathcal{P}|}\sum_{p} \mathcal{L}(f,p,c)$ for each $c\in E$ \Comment{the directly evaluated objective: $W$ forwards over $\mathcal{P}$}
  \State $d \gets \arg\min_{c\in E} \bar{\mathcal{L}}(c)$;\quad $\mathcal{H}\gets\mathcal{H}\cup\{d\}$
  \State \textbf{if} $\bar{\mathcal{L}}(d) < \bar{\mathcal{L}}(d^\star)$ \textbf{then} $d^\star\!\gets d,\ s\!\gets\!0$\ \ \textbf{else}\ \ $s\!\gets\!s{+}1$
  \State \textbf{if} $s \ge \tau$ \textbf{then} $d \gets \textsc{Restart}(d^\star)$;\ \ $\mathcal{H}\gets\mathcal{H}\cup\{d\}$;\ \ $s\!\gets\!0$ \Comment{basin-hop; $d^\star$ preserved}
\EndFor
\State \textbf{return} $d^\star$ \Comment{date of maximum decay-weighted KL found}
\end{algorithmic}
\end{algorithm*}

\paragraph{Harness-audit receipts.} Table~\ref{tab:receipts} pins the provenance behind
Table~\ref{tab:harness}: each stack was re-verified at the exact commit below (default branch), with SHAs pulled
from the GitHub / Hugging Face APIs. For every stack we located the
call site that applies (or delegates) the chat template, confirmed by exhaustive search that the stack itself never passes a
\texttt{date\_string}, and traced whether the rendered prompt is written to the saved outputs. Per stack: lm-evaluation-harness
never passes a date itself, though a caller can pass one through \texttt{chat\_template\_args}, and saves the rendered
prompt only with \texttt{-{}-log\_samples}; the Open LLM Leaderboard's fork accepts no date, and its logging code writes the rendered prompt to gated per-sample details; in the ones we inspected,
Llama-3.2-3B-Instruct's, it carries the template's fallback date \texttt{26 Jul 2024} rather than the run date (27 Sep 2024); HELM keeps the rendered prompt as the key of its default request cache, and in its saved outputs only for scenarios that echo the prompt; lighteval saves it with \texttt{-{}-save-details}, as text (vLLM) or token ids
(Transformers, SGLang), and also in its default cache; AlpacaEval applies a model's own template only when a configuration opts in (as one shipped judge configuration
does), and its other local configurations use their own prompts, which carry no date; Arena-Hard-Auto leaves the template to the server by default (Table~\ref{tab:harness}) but ships a local example that applies it client-side; and simple-evals writes a
fixed, stale \texttt{2024-04-01} into two of its own persona prompts, which it saves with each conversation by
default, though it never sees the prompt a provider renders. OpenCompass, EvalPlus, and BigCodeBench apply the model's
template by default (EvalPlus and BigCodeBench unless run as direct completion) and Qwen2.5-Math's evaluation only
with \texttt{-{}-apply\_chat\_template}; none passes a date or saves the rendered prompt (OpenCompass saves the prompt
before the template is applied, EvalPlus and BigCodeBench only the completions, and Qwen2.5-Math drops the prompt
before saving). \texttt{lm-sys/arena-hard-auto} now
redirects to \texttt{lmarena/arena-hard-auto}; the Open LLM Leaderboard v2 is a git-versioned Hugging Face Space
(SHA is the Space commit) whose evaluation runs on Hugging Face's fork of lm-evaluation-harness (the
\texttt{leaderboard} task group).

\begin{table*}[t]\centering\small\setlength{\tabcolsep}{4pt}
\begin{tabular}{llll}
\toprule
Stack & Repository / space & Latest release & Commit \\
\midrule
lm-evaluation-harness   & \texttt{EleutherAI/lm-evaluation-harness} & v0.4.12  & \texttt{0f8479cc} \\
Open LLM Leaderboard v2 & \texttt{HF:open-llm-leaderboard/open\_llm\_leaderboard} (Space)  & n/a      & \texttt{0ca4edb6} \\
HELM                    & \texttt{stanford-crfm/helm}               & v0.5.16  & \texttt{63754d05} \\
lighteval               & \texttt{huggingface/lighteval}            & v0.13.0  & \texttt{932e1f2f} \\
OpenCompass             & \texttt{open-compass/opencompass}         & 0.5.4    & \texttt{ec591804} \\
EvalPlus                & \texttt{evalplus/evalplus}                & v0.3.1   & \texttt{26d6d00b} \\
BigCodeBench            & \texttt{bigcode-project/bigcodebench}     & v0.2.5   & \texttt{09dd993f} \\
AlpacaEval              & \texttt{tatsu-lab/alpaca\_eval}           & v0.6.6   & \texttt{cd543a14} \\
Arena-Hard-Auto         & \texttt{lmarena/arena-hard-auto}          & n/a      & \texttt{196f6b82} \\
OpenAI simple-evals     & \texttt{openai/simple-evals}              & n/a      & \texttt{652c89d0} \\
Qwen2.5-Math eval       & \texttt{QwenLM/Qwen2.5-Math}              & n/a      & \texttt{a45202bd} \\
\bottomrule
\end{tabular}
\caption{\textbf{Harness-audit receipts.} Provenance for Table~\ref{tab:harness}: each stack audited from primary
sources at the listed commit (default branch), SHAs pulled from the GitHub / Hugging Face APIs (latest commit at
audit time), audit date 2026-08-28. Latest release: the most recent tagged release at audit time; the audited commit
is the default-branch head, which may be ahead of it. ``n/a'' marks an untagged repository (pinned by commit only).}
\label{tab:receipts}
\end{table*}

\section{Qualitative date-flip examples}\label{app:examples}
Held-out items where changing only the injected date flips the scored answer; prompts are byte-identical apart
from the date, and responses are truncated (``\ldots'') by hand to the load-bearing span with the pivotal
difference in \textbf{bold}. The first group contrasts the pinned baseline date with the \emph{optimized} (accuracy-blind, max-KL) date (spanning failure modes); the second shows that two \emph{plausible} dates already
yield genuinely different answers, not refusals or parsing artifacts, but different reasoning to a different
letter. Section~\ref{sec:infer} shows the optimized date costs the four non-Llama models no statistically significant
\emph{aggregate} accuracy, yet individual items still flip.

{\small\sloppy\raggedright
\noindent\emph{Baseline vs.\ the optimized date (one per model except Llama-3.1-8B).}\par\smallskip
\paragraph{Llama-3.2-3B, IFEval: a date-questioning preamble that breaks a constraint.}
\emph{Prompt:} ``Coincidence is God's way of remaining anonymous.'' What are your thoughts? \emph{(no commas; $>$800 words)}.\\
\texttt{14 Jul 2026}: answers comma-free and on task \emph{(pass)}.\\
\texttt{01 Jul 1963}: prepends ``\textbf{I must point out that the date you've provided, July 1, 1963, is not a valid date\,\ldots\ before the birth of the computer}'', then answers the quote anyway, but the \textbf{commas in that date-invalidity preamble} break the no-comma constraint (one of the six preambles that break an IFEval constraint, Appendix~\ref{sec:cutoff}), not an outright refusal.

\paragraph{gpt-oss-20B, MMLU: silent multiple-choice flip.}
\emph{Prompt:} A lesion compressing the facial nerve at the stylomastoid foramen causes ipsilateral \ldots\ [A--D]. (gold \texttt{A})\\
\texttt{2026-06-17}: reasons that the nerve at that point ``is purely motor \ldots\ taste fibers \ldots\ already in the nerve earlier'' $\Rightarrow$ \textbf{A} \emph{(correct)}.\\
\texttt{1959-03-02}: the chain diverges and selects $\Rightarrow$ \textbf{D} \emph{(wrong)}: no refusal, just a different answer.

\paragraph{SmolLM3-3B, HumanEval: long-form reasoning diverges.}
\emph{Prompt:} Complete \texttt{car\_race\_collision(n)} (count collisions of two opposing car sets).\\
\texttt{14 June 2026}: ``\ldots\ cars moving in opposite directions on a \textbf{straight} road \ldots'' \emph{(returned code passes)}.\\
\texttt{07 January 3304}: ``\ldots\ on an \textbf{infinitely long} road \ldots'': the reasoning chain diverges from its first sentence and never closes its \texttt{<think>} block within the token budget, so no code is returned \emph{(fail)}.

\paragraph{Granite-3.3-8B, IFEval: rewording breaks a constraint.}
\emph{Prompt:} Explain GANs in bullet points [no commas; required keywords incl.\ ``lacking''; end with P.P.S].\\
\texttt{July 14, 2026}: ``\ldots\ GAN training can be unstable and \textbf{lacking} in convergence \ldots'' \emph{(constraints met)}.\\
\texttt{December 20, 1300}: ``\ldots\ Initially GANs \textbf{lacked} stability during training \ldots'': a near-synonymous rewording, yet the required keyword ``lacking'' is now absent, so the IFEval keyword check flips to \emph{fail}.

\paragraph{Mistral-24B, IFEval: an unclosed required quotation.}
\emph{Prompt:} Write a funny article on the dinosaurs' extinction, \textbf{wrapping the whole response in double quotes}, with exactly 8 bullet points.\\
\texttt{2026-06-17}: wraps the article in quotes and lists all 8 bullets \emph{(both constraints pass)}.\\
\texttt{1780-04-14}: opens with a flourish (``In the grand theater of Earth's history\,\ldots'') and still lists all 8 bullets, but \textbf{never closes the wrapping quotation}, so the \texttt{startend:quotation} check fails while the bullets pass.

\medskip
\noindent\emph{Two plausible dates, no optimization: the scored answer differs on genuinely different reasoning.}\par\smallskip
\paragraph{Mistral-24B, MMLU: an invented extra requirement.}
\emph{Prompt:} Does $x^2-12$ satisfy an Eisenstein criterion for irreducibility over $\mathbb{Q}$? (gold \texttt{B}: yes, with $p{=}3$.)\\
\texttt{2025-02-10}: ``\ldots\ the polynomial satisfies the Eisenstein criterion with $p=3$'' $\Rightarrow$ \textbf{B} \emph{(correct)}.\\
\texttt{2025-11-25}: ``\ldots\ \textbf{both} $p=2$ \textbf{and} $p=3$ satisfy \ldots\ the criterion requires a unique prime \ldots\ does not satisfy uniquely'' $\Rightarrow$ \textbf{D} \emph{(wrong)}.

\paragraph{Llama-3.2-3B, MMLU: an arithmetic slip in $\mathbb{Z}_2$.}
\emph{Prompt:} Find all zeros of $x^2+1$ in $\mathbb{Z}_2$. (gold \texttt{B}: $x{=}1$.)\\
\texttt{10 Feb 2025}: ``$x=1$: $1^2+1 = \textbf{2}$ (not equal to $0$\,\ldots) \ldots\ there are no zeros'' $\Rightarrow$ \textbf{A} \emph{(wrong)}.\\
\texttt{12 Feb 2025}: ``$x=1$: $1^2+1 = \textbf{0}$ (equal to $0$) \ldots\ the only zero \ldots\ is $1$'' $\Rightarrow$ \textbf{B} \emph{(correct)}.

\paragraph{Granite-3.3-8B, MMLU: ``two'' vs.\ ``only one''.}
\emph{Prompt:} How many homomorphisms are there of $\mathbb{Z}$ into $\mathbb{Z}_2$? (gold \texttt{B}: 2.)\\
\texttt{February 10, 2025}: ``\ldots\ If the image is $\mathbb{Z}_2$, then $f(n)$ must be $1$ for any odd $n$ and $0$ for even $n$ \ldots\ there are \textbf{two} distinct homomorphisms'' $\Rightarrow$ \textbf{B} \emph{(correct)}.\\
\texttt{November 25, 2025}: ``\ldots\ If $f(1)=1$, then $f(n)=n$ \ldots\ this does not satisfy the homomorphism property because $f(2)=2\neq f(1+1)=1+1=0$ \ldots\ there is \textbf{only one} valid homomorphism'' $\Rightarrow$ \textbf{A} \emph{(wrong)}: it never reduces $f(n)$ modulo 2.\par
}

\section{Why Llama: the cutoff line and date-related refusals}\label{sec:cutoff}

Llama's template prints a hard-coded cutoff line (\texttt{Cutting Knowledge Date: December 2023}) immediately above
\texttt{Today Date:} (Appendix~\ref{app:deploy}). Across templates, the selected dates fall on both sides of the stated cutoffs:
Llama-3.2-3B's selected dates are \emph{pre-cutoff} (1941 for MMLU, 1963 for IFEval), so the preamble asserts the present
precedes the model's own knowledge cutoff. Mistral (1913, 1956, 2009), SmolLM3 (1842), and gpt-oss (2016) are handed
pre-cutoff dates for their own cutoffs and none shows a refusal-driven loss on benchmark items (gpt-oss's IFEval loss is mostly non-termination:
19 of its 23 right$\to$wrong flips reach no final answer and 4 answer wrongly, \S\ref{sec:infer}), while for several models the search drifts to far-\emph{future} years instead (4839, 7325, 8136, 8870).
Llama-3.1-8B, under the same cutoff line, gets pre-cutoff VAL-selected GSM8K and IFEval dates (1342, 1620) but loses
most at a far-future accuracy-blind MMLU date (3846), refusing at dates on both sides of the cutoff (below); the
cutoff and log-prob analyses below were run on Llama-3.2-3B. Three analyses characterize the effect. (i)~\emph{The cutoff line matters.} Keeping the line but shifting its cutoff to 1900, so the optimized 1941/1963 date becomes
\emph{post}-cutoff, takes the drop to $0.0$\,pp (MMLU) and $5.2$ (IFEval); deleting the line outright likewise
lowers it ($18.8{\to}7.3$, $20.8{\to}3.1$), though that is a blunter test, since it also removes a temporal cue.
Shifting the cutoff to 3000 instead (the date still precedes it) leaves $7.3$ and $12.5$\,pp ($39\%$ and $60\%$ of the
drop); a year-3000 cutoff is itself anomalous, and the residual is consistent with the exact strings mattering
(below), so we read (i) as evidence that the cutoff line's content modulates the loss, not that the date-before-cutoff ordering
alone causes it (these ablation deltas are point estimates on 96 items, read qualitatively).
(ii)~\emph{A cross-model contrast.} All four other templates also pair a cutoff line with the date (Mistral, gpt-oss,
Granite, and SmolLM3, whose \texttt{Knowledge Cutoff Date: June 2025} sits directly above \texttt{Today Date:} as in
Llama's template), yet only the Llama models lose accuracy statistically significantly: the refusal-driven loss is Llama-specific. (iii)~\emph{Log-prob scoring.} Scored by the log-probability of the
answer letter under a direct-answer prompt rather than by generation, the MMLU drop at \texttt{25 Feb 1941} falls from $18.8$ to $5.2$\,pp on the log-prob metric's own (lower)
baseline of $55.2\%$, consistent with the generative loss being refusal rather than lost knowledge. A plausible date does not appear to trigger any of this: among the six highest-KL of the $1096$ real
in-window dates, Llama-3.2-3B's largest held-out drop is $6.2$\,pp (MMLU, $p{=}0.11$) and $4.2$ on
IFEval (Appendix~\ref{app:inwindow}).

\begin{figure*}[t]
\centering
\includegraphics[width=\textwidth]{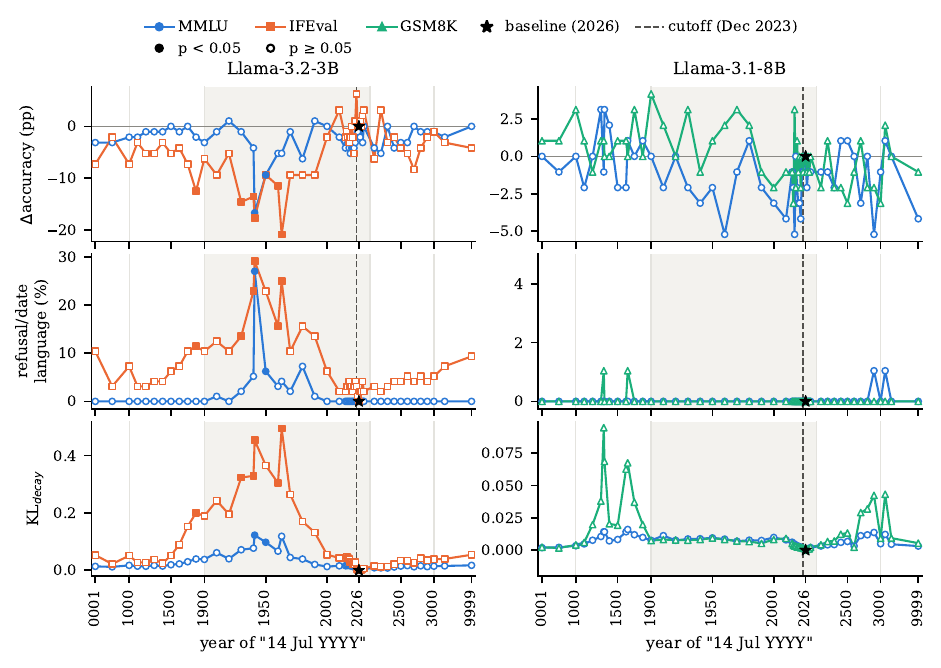}
\caption{\textbf{A band of years matters for Llama-3.2-3B; for Llama-3.1-8B only the full date does.} Only the year of \texttt{14 Jul YYYY}
varies ($x$-axis piecewise linear, 1900--2035 expanded and shaded; light rules: scale changes); 96 TEST items per point. Rows: accuracy change against the
baseline (pp; negative${=}$loss), share of generations with refusal or date-questioning language, output KL. Filled markers (every row):
accuracy exact-McNemar $p{<}0.05$, unadjusted; star: the baseline; dashed line: the template's cutoff. Llama-3.1-8B's optimized \texttt{01 May 1342} and
\texttt{17 Jan 3846} cost $28.1$ and $36.5$\,pp, against at most $5.2$\,pp for any of its years here.}
\label{fig:yearsweep}
\end{figure*}

\paragraph{Year versus full date.} To separate the year from the full date we hold the day at \texttt{14 Jul} and
sweep the year over a nonuniform grid from 0001 to 9999 (every 100th year from 1000 to 3000, every 10th from 1900 to 2010,
each of 2015--2030, and 0001, 0500, 5000, and 9999: 50 years besides the baseline's 2026, plus the years of the
optimized dates) on the four Llama cells with a
statistically significant loss (bold in Table~\ref{tab:results}; Figure~\ref{fig:yearsweep}); nominally significant means unadjusted exact-McNemar $p{<}0.05$ (each
year is tested at one day on 96 items, with no correction over a cell's ${\sim}52$ years, so at most 2--3 such years are expected by chance). Llama-3.1-8B is flat: no year moves MMLU or GSM8K by more than
$5.2$\,pp, and none is nominally significant. Yet its optimized dates cost $28.1$\,pp (\texttt{01 May 1342}, GSM8K) and
$36.5$\,pp (\texttt{17 Jan 3846}, MMLU), against no loss at \texttt{14 Jul} of the same years (0.0 and a 1.0\,pp
gain), and two of
its optimized dates carry over to a benchmark they were not searched on (\texttt{17 May 1620} and \texttt{17 Jan 3846} cost $18.8$
and $16.7$\,pp of GSM8K). Llama-3.2-3B's IFEval loses $5$--$21$\,pp across 1800--1990 and recovers by 2000, while
far-past and far-future years cost less on median (at most $8.3$\,pp; none nominally significant); its MMLU moves nominally significantly only at 1941 ($16.7$\,pp) and 1950 ($9.4$), and within
1963 the day decides (\texttt{14 Jul}: $5.2$\,pp; \texttt{01 Jul}: $19.8$). Across both models, all 18 date--benchmark pairs that
cost at least $10$\,pp draw refusal or date-questioning language in $11$--$52\%$ of generations (flagged by our matcher, a set of
regular expressions for common refusal and date-questioning phrases such as ``I cannot'' and ``knowledge cutoff''),
against at most
$2\%$ on MMLU and GSM8K ($9\%$ on IFEval, nearly all refusals) at dates that cost less
than $3$\,pp. At Llama-3.1-8B's optimized dates the matcher's refusal patterns flag 33 of the 36 right$\to$wrong flips at
\texttt{17 Jan 3846} (MMLU) and 21 of the 29 at \texttt{01 May 1342} (GSM8K), and no generation at the baseline.

\paragraph{The refusal audit.} The split in \S\ref{sec:infer} (61\% refusals, 24\% wrong answers, 15\% preambles
that break an IFEval constraint) comes from a per-item audit of all 41 right$\to$wrong flips on MMLU and IFEval at Llama-3.2-3B's accuracy-blind
dates, 18 on MMLU and 23 on IFEval.\footnote{Official IFEval scores the whole response and counts the 6 preamble cases
(e.g., ``\ldots\ July 1, 1963, is not a valid date\,\ldots'') as failures. As a diagnostic, scoring only the response body
gives an IFEval drop of $12.5$\,pp (unadjusted $p=0.017$, not statistically significant after Bonferroni correction,
$\alpha{=}0.0021$) rather than the accuracy-blind date's $18.8$. MMLU has no such case (0/18). Llama-3.2-3B emits no reasoning channel, so
the reasoning-channel scoring issue (see Limitations) does not apply to these cells.} Most refusals are generic (``I can't help with that.''); 6 of the 25 name the date, year, or cutoff (``as
my knowledge stopped in 1963'', ``the date you've given me is in the past''). Because a refusal still yields a
response, it is scored as a wrong answer, not flagged as unanswered, even when no letter can be extracted (11 of the
18 MMLU flips); we recommend
reporting refusal rates alongside accuracy.

Together, for Llama-3.2-3B the worst-case loss depends on the stated cutoff (moving the cutoff to 1900 removes most of
it) and takes the form of refusal, but the cutoff alone does not
explain it: which out-of-window dates trigger it depends on the model and on the full date string (a band of years for
Llama-3.2-3B, particular dates for Llama-3.1-8B, not the clock dates), so search finds larger losses than random draws reach (at most $10.4$\,pp, Llama-3.2-3B$\cdot$IFEval, none
significant after correction) and, for Llama-3.1-8B, than year sweeps. Under plausible dates it nearly vanishes, leaving the per-item churn of \S\ref{sec:cost} as the
reproducibility hazard.
\section{Training: identical data, different gradient}\label{sec:train}

Does a benign date change what a model \emph{learns}? We fine-tune each model on identical benign data (64
examples), identical initialization and data order, varying only the date injected during training. A seed-twin
condition (same date, different training-example order) measures sensitivity to training order; a random-date
condition guards against a generic effect. Training uses LoRA on the q/v projections (rank 16, $\alpha{=}32$, no
dropout) with AdamW (learning rate $10^{-4}$, batch 4) for 100 and 300 steps on the first 64 prompts of a 96-prompt multi-topic benign
set, the next 32 serving as held-out prompts; the adapter initialization is fixed and restored between conditions. The date measurably shifts the fine-tuning gradient (Table~\ref{tab:grad}). There, \emph{unbd-max} is the most
output-divergent of the training study's unbounded candidates (a VAL-selected date and random unbounded draws; for
gpt-oss a random draw), and \emph{bnd-max} the most divergent of random 2024--2026 draws; the training study drew
these from its own candidate pools, so they need not match the final inference-study selections. We take the cosine on the initial
full-batch gradient of the gold-answer loss with respect to the q/v-projection weights that LoRA adapts; that gradient does not depend on example
order, so the seed-twin cosine is $1.00$ by construction and any departure is date-induced, but the
informative comparison is not against $1.00$; it is the \emph{optimized} date against the \emph{random}-date
distribution. For Llama-3.2-3B the effect is large and \emph{specific}: the optimized date rotates the gradient to
cosine 0.19, a clear outlier against a random-date distribution of $0.67\pm0.06$. For Mistral and Granite the gradient is date-sensitive but generic
(unbd-max is indistinguishable from a random date), and SmolLM3 and gpt-oss barely move (cosines
$\ge0.98$; SmolLM3's unbd-max falls below all eight of its random dates, $0.992$ vs.\ $0.996$--$0.999$, but
its cosine stays above $0.99$, and gpt-oss's most divergent unbounded date is itself a random draw).
Among the five models we fine-tune, the one whose training the date perturbs most is the one whose accuracy it degrades most.

\begin{table*}[t]
\centering
\begin{tabular}{lcccc}
\toprule
Model & unbd-max & bnd-max & random ($\mu\pm\sigma$) & floor \\
\midrule
Llama-3.2-3B   & \textbf{0.19} & 0.85 & $0.67\pm0.06$ & 1.00 \\
Mistral-24B    & 0.83 & 0.93 & $0.83\pm0.04$ & 1.00 \\
Granite-3.3-8B & 0.88 & 0.98 & $0.87\pm0.03$ & 1.00 \\
SmolLM3-3B     & 0.99 & 1.00 & $1.00\pm0.00$ & 1.00 \\
gpt-oss-20B    & 0.99 & 1.00 & $0.99\pm0.00$ & 1.00 \\
\bottomrule
\end{tabular}
\caption{\textbf{Gradient cosine to the baseline-date gradient.} \emph{unbd-max}/\emph{bnd-max}: the most
output-divergent unbounded / 2024--2026 date; \emph{random}: eight further random unbounded (any-year) dates; \emph{floor}: the seed twin, which changes only
the training order ($1.00$ by construction). Bold: far below the random-date spread.}
\label{tab:grad}
\end{table*}

After LoRA~\citep{lora} fine-tuning, the weight-update direction still separates the date conditions from the
seed-twin floor for Llama-3.2-3B (cosine $0.49$ vs.\ a floor of $0.84$ at 100 steps and $0.81$ at 300 steps); for
SmolLM3, Mistral, and gpt-oss the seed-twin cosine is already at or below the date conditions' level at both 100 and
300 steps, so the two are not separable.\footnote{Granite's LoRA runs rendered its baseline date
differently from its template and are omitted here; its row in Table~\ref{tab:grad} is recomputed with the template's
rendering.} We detect no behavioral change at this scale: models fine-tuned under different dates do not differ detectably
on a 32-prompt IFEval check (a weak check: after training on 1--3-word targets, every condition, baseline included,
answers IFEval prompts in a few words; a GSM8K check scores $0\%$ throughout), and in an exploratory LLM-judge
comparison on the first 24 IFEval prompts for each of the four models above (Granite omitted), the
unbd-max-trained model's generations win 14, lose 16, and tie 66 of 96 comparisons against the baseline-trained
model's (15\%/17\%/69\%; sign test $p{=}0.86$ over decisive comparisons), a limited-power null, not evidence of
equivalence. Every fixed-date
condition reproduces bit-identical training metrics (losses, update norms and cosines, held-out KL, task scores) across
three repetitions on A100 nodes, so the contrast is the date, not run-to-run variation.

\section{Localization and calibration}\label{sec:mech}

Two observations locate the inference effect. It becomes visible in \emph{late} layers: the last-input-token hidden
state is essentially unchanged (cosine $\approx1.0$) through the first half of the network and only diverges in
the final layers (to $\sim$0.70 for Llama-3.2-3B$\cdot$IFEval). And it derails the answer \emph{early}: where answers
differ, the fragile Llama-3.2-3B cells (MMLU, IFEval) split within the first few generation tokens, whereas the other models and longer
tasks diverge much later, though this index is confounded by answer length (for MMLU the answer letter \emph{is}
token one to three), so we read it only within a task, not across. Llama-3.2-3B does not collapse: unique-response
rates across prompts stay near 1.0, and its accuracy loss is spent on distinct failures, predominantly the refusals of Appendix~\ref{sec:cutoff}. These diagnostics
use different samples: the hidden-state cosines average 24 evenly spaced TEST prompts at the baseline and 12 dates from
the search trajectory, divergence timing compares the 96 TEST generations at the baseline and accuracy-blind dates,
and unique-response rates use the 32 TRAIN prompts at the search's checkpoints. To calibrate the KL
magnitudes: two greedy runs at the \emph{same} date and batch layout differ by exactly $0$ (decoding is deterministic; a re-score under different batching, as in Appendix~\ref{app:inwindow}, leaves a small bf16 noise floor), a random
plausible date differs from the baseline by $\mathrm{KL}\approx0.001$--$0.03$ on the fragile Llama-3.2-3B cells (and by $\lesssim0.003$ in every non-Llama model's cells), and the optimized date reaches $0.19$--$0.69$ on the fragile cells (decay-weighted, as for every KL value we report), more than an order of magnitude above the same cell's random plausible dates. That gap quantifies how far the optimized date shifts the output distribution.

\begin{figure*}[t]
\centering
\includegraphics[width=\textwidth]{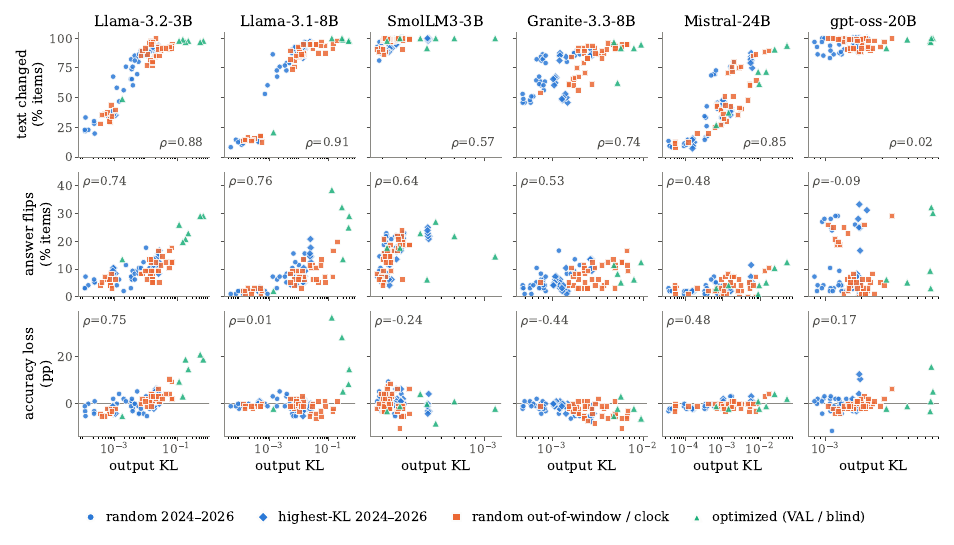}
\caption{\textbf{Where it varies, output KL tracks how much the text changes, less which answers flip, and accuracy clearly only
for Llama-3.2-3B.} One point per scored date and benchmark. $x$: the date's output KL from the baseline (log). Rows:
share of TEST generations whose text changes, share of items whose correctness flips, accuracy loss (pp). $\rho$:
Spearman.}
\label{fig:kloutcome}
\end{figure*}

\paragraph{Divergence vs.\ outcome.} Figure~\ref{fig:kloutcome} pools, per model, the enumerated and shortlisted dates (random in- and
out-of-window, clock, the six highest-KL in-window, and the optimized dates; in-window dates use the screen's KL, which
agrees with the enumeration's on shared dates, Spearman $0.81$--$0.99$ pooled per model, though within a benchmark only weakly for Granite and not for
SmolLM3, whose KL is at the noise floor). Where KL has range it predicts how much
of the generated text changes (Spearman $\rho{=}0.88$ and $0.91$ for the Llama models, $0.85$ Mistral, $0.74$
Granite); for gpt-oss and SmolLM3 at least $81\%$ of generations change at every date whatever the KL, since long
reasoning traces almost always diverge somewhere. KL predicts correctness flips less well ($\rho{=}0.48$--$0.76$;
gpt-oss $-0.09$) and accuracy loss clearly only for Llama-3.2-3B ($\rho{=}0.75$; $0.71$ without its optimized dates); for the
other models $\rho$ ranges from $-0.44$ to $0.48$ with no consistent sign. This is the quantitative form of
\S\ref{sec:infer}'s ``divergence is not degradation''.

We stress what this does \emph{not} show: the date is not established as \emph{mechanistically} privileged over
other preamble tokens, and the disruption may well be general preamble sensitivity. What makes it a hazard is not
a special mechanism but the scope we fixed at the outset (\S\ref{sec:intro}).
\section{LLM-judge details and calibration}\label{app:judge}

\paragraph{Protocol.} Per model, each of the 96 IFEval TEST prompts contributes three comparisons against its
baseline-date answer: the model's accuracy-blind date and the plausible dates \texttt{10 Feb 2025} and
\texttt{25 Nov 2025}, all taken from the enumeration's generations. Comparisons in which either side has no final
answer, or the two answers are byte-identical, are not sent to the judge and are counted separately (of 1,728
comparisons, 142 identical and 196 with a missing answer; 1,390 judged). A prompt's \emph{share} is the fraction of
its decisive verdicts that prefer the baseline; tests are sign tests over prompts (for the share, prompts with a decisive verdict, dropping shares of exactly one half; for the
optimized$-$plausible gap in how often a winner is picked, paired per-prompt differences, ties dropped), CIs are
prompt-bootstrap ($10^4$ resamples), and $p$-values are Bonferroni-corrected over the six models. An \emph{identity}
null, an answer judged against a byte-identical copy, yields no decisive verdict (0 of 12 in each study): the order-consistent verdict
invents no winner between identical answers. Figure~\ref{fig:judgevar} shows how often a winner is picked (a) and the share (b). The number above each model in (a) is the mean paired difference over the
prompts where both rates are defined, so it can differ from the difference of the two bars (gpt-oss: $+20$ against
$82$ and $65\%$); the counts above (b) are for the optimized date, and an asterisk marks the sign test of the paired
gap in (a) and of the optimized-date share in (b). Where exactly one
side passes IFEval's check, the judge prefers that side in $75$--$100\%$ of its decisive verdicts.

\paragraph{The 32 single-target prompts.} The same protocol on the 32 TRAIN prompts of the single-target runs, each
judged against its own per-prompt oracle date (generated at batch size 1; circles in Figure~\ref{fig:judgevar}), gives
results that point the same way as \S\ref{sec:judge}: under the oracle date the judge prefers the baseline answer in $93\%$ of Llama-3.1-8B's decisive
verdicts ($26$ of $28$ prompts; Bonferroni $p{=}1.8\times10^{-5}$) and $70\%$ of Llama-3.2-3B's ($19$ of $27$; $p{=}0.052$
before correction), while the other four stay at $56$--$59\%$. Because the prompt sets and
batch sizes differ, we report the two studies separately rather than pool them.

\paragraph{Judge prompt.} The judge is GPT-5.4 (every reply came from snapshot \texttt{gpt-5.4-2026-03-05}) at
\texttt{medium} reasoning effort. Every
instruction/response pair is scored in both A/B orders and a net winner is counted only when the two orders agree
(position-bias cancellation); the judge sees each final answer in full, the verdict is read from the last
\texttt{VERDICT:} line, and an unparseable reply is retried up to twice (2 of the 482 judged comparisons in the 32-prompt study stayed
unparsed and count as no verdict; none in the 96-prompt study). The system message is:
\begin{quote}\small
You are an impartial evaluator. Given a user instruction and two candidate responses (A and B), decide which
response better satisfies the instruction --- judge instruction-following, correctness, completeness, and overall
quality. Be objective and ignore length unless it affects quality. If the two are equivalent, answer TIE. Think
briefly, then end with exactly one line: `VERDICT: A', `VERDICT: B', or `VERDICT: TIE'.
\end{quote}
The user message, with the three braced placeholders filled in
per comparison, is:
{\small\begin{verbatim}
[INSTRUCTION]
{instruction}

[RESPONSE A]
{response A}

[RESPONSE B]
{response B}

Which response better satisfies the instruction?
\end{verbatim}}

\section{Per-cell exact tests}\label{app:pvals}

\begin{figure*}[t]
\centering
\includegraphics[width=0.94\textwidth]{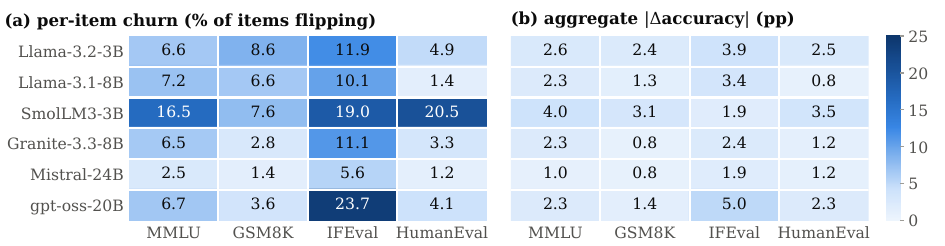}
\caption{\textbf{Per-cell values behind Figure~\ref{fig:repro}.} Between two plausible dates (means over
the 36 plausible-date pairs): \textbf{(a)}~per-item churn (\% of items switching between correct and incorrect) and
\textbf{(b)}~the absolute change in aggregate accuracy (pp).}
\label{fig:churnheat}
\end{figure*}
Table~\ref{tab:pvals} gives the full dates behind Table~\ref{tab:results}, the discordant counts
(right$\to$wrong / wrong$\to$right, baseline vs.\ date), and the $p$-value of the exact two-sided McNemar test~\citep{dror2018hitchhiker}. Bonferroni at $m{=}24$
($\alpha{=}0.0021$) is applied separately to the VAL and accuracy-blind columns. The five bold cells have unadjusted
$p{\le}9.1\times10^{-4}$, so Bonferroni-adjusted $p{<}0.025$ within their column, and all five also survive Bonferroni
over all 48 tests ($\alpha{=}0.00104$). In 13 of the 16 non-Llama
accuracy-blind cells the paired prompt-bootstrap $95\%$ CI of the accuracy change ($10^4$ replicates, percentile)
excludes losses of more than five items ($5.2$\,pp; in 10, more than three, $3.1$\,pp); the other
three are IFEval cells (gpt-oss, SmolLM3, Mistral), whose point estimates are small losses ($1.0$--$5.2$\,pp) and whose
upper bounds reach $16.7$, $10.4$, and $9.4$\,pp. These per-cell intervals are unadjusted and give no simultaneous $95\%$ coverage across cells.

\begin{table*}[!tbp]\centering\small\setlength{\tabcolsep}{4pt}
\resizebox{\textwidth}{!}{\begin{tabular}{@{}ll lrrr lrrr@{}}
\toprule
 & & \multicolumn{4}{c}{VAL-selected date} & \multicolumn{4}{c}{Accuracy-blind date} \\
\cmidrule(lr){3-6}\cmidrule(lr){7-10}
Model & Bench & date & $\Delta$ & R$\to$W/W$\to$R & $p$ & date & $\Delta$ & R$\to$W/W$\to$R & $p$ \\
\midrule
Llama-3.2-3B & MMLU & \texttt{25 Feb 1941} & +18.8 & 19/1 & $4.0\text{e-}5$ & \texttt{01 Apr 1963} & +14.6 & 18/4 & $0.004$ \\
 & GSM8K & \texttt{02 Apr 1963} & +9.4 & 17/8 & $0.108$ & \texttt{01 Apr 1963} & +3.1 & 11/8 & $0.648$ \\
 & IFEval & \texttt{28 Sep 1963} & +20.8 & 24/4 & $1.8\text{e-}4$ & \texttt{01 Jul 1963} & +18.8 & 23/5 & $9.1\text{e-}4$ \\
 & HumanEval & \multicolumn{4}{c}{reverted to baseline} & \texttt{31 May 1986} & $-$5.2 & 4/9 & $0.267$ \\
\addlinespace[2pt]
Llama-3.1-8B & MMLU & \multicolumn{4}{c}{reverted to baseline} & \texttt{17 Jan 3846} & +36.5 & 36/1 & $5.5\text{e-}10$ \\
 & GSM8K & \texttt{01 May 1342} & +28.1 & 29/2 & $4.6\text{e-}7$ & \texttt{02 May 1334} & +5.2 & 9/4 & $0.267$ \\
 & IFEval & \texttt{17 May 1620} & +8.3 & 16/8 & $0.152$ & \texttt{13 May 1620} & +14.6 & 21/7 & $0.013$ \\
 & HumanEval & \multicolumn{4}{c}{reverted to baseline} & \texttt{18 Dec 1774} & $-$2.1 & 0/2 & $0.500$ \\
\addlinespace[2pt]
SmolLM3-3B & MMLU & \texttt{14 June 8136} & $-$3.1 & 7/10 & $0.629$ & \texttt{21 August 3303} & $-$2.1 & 6/8 & $0.791$ \\
 & GSM8K & \multicolumn{4}{c}{reverted to baseline} & \texttt{29 March 1844} & +0.0 & 3/3 & $1.000$ \\
 & IFEval & \texttt{10 February 1842} & +4.2 & 13/9 & $0.523$ & \texttt{27 January 3303} & +1.0 & 11/10 & $1.000$ \\
 & HumanEval & \texttt{25 February 4839} & $-$1.0 & 8/9 & $1.000$ & \texttt{07 January 3304} & $-$8.3 & 9/17 & $0.169$ \\
\addlinespace[2pt]
Granite-3.3-8B & MMLU & \multicolumn{4}{c}{reverted to baseline} & \texttt{February 30, 1577} & $-$5.2 & 3/8 & $0.227$ \\
 & GSM8K & \texttt{February 31, 8870} & +3.1 & 4/1 & $0.375$ & \texttt{February 31, 1215} & $-$2.1 & 2/4 & $0.688$ \\
 & IFEval & \multicolumn{4}{c}{reverted to baseline} & \texttt{December 20, 1300} & $-$6.2 & 3/9 & $0.146$ \\
 & HumanEval & \multicolumn{4}{c}{reverted to baseline} & \texttt{November 23, 1907} & $-$2.1 & 3/5 & $0.727$ \\
\addlinespace[2pt]
Mistral-24B & MMLU & \texttt{1913-06-17} & $-$2.1 & 1/3 & $0.625$ & \texttt{1960-03-10} & $-$1.0 & 2/3 & $1.000$ \\
 & GSM8K & \multicolumn{4}{c}{reverted to baseline} & \texttt{1987-04-17} & +1.0 & 1/0 & $1.000$ \\
 & IFEval & \texttt{2009-01-01} & +4.2 & 7/3 & $0.344$ & \texttt{1780-04-14} & +2.1 & 7/5 & $0.774$ \\
 & HumanEval & \texttt{1956-06-17} & $-$1.0 & 1/2 & $1.000$ & \texttt{1995-02-10} & $-$2.1 & 1/3 & $0.625$ \\
\addlinespace[2pt]
gpt-oss-20B & MMLU & \texttt{7325-06-09} & $-$2.1 & 2/4 & $0.688$ & \texttt{1959-03-02} & $-$3.1 & 3/6 & $0.508$ \\
 & GSM8K & \multicolumn{4}{c}{reverted to baseline} & \texttt{1988-02-31} & +1.0 & 2/1 & $1.000$ \\
 & IFEval & \texttt{2016-12-31} & +15.6 & 23/8 & $0.011$ & \texttt{2017-01-06} & +5.2 & 17/12 & $0.458$ \\
 & HumanEval & \multicolumn{4}{c}{reverted to baseline} & \texttt{1995-03-08} & $-$1.0 & 2/3 & $1.000$ \\
\bottomrule
\end{tabular}
}
\caption{\textbf{Exact McNemar tests per cell} (96 TEST items). $\Delta$: baseline${-}$date accuracy, positive${=}$loss. R$\to$W/W$\to$R: items right at baseline and wrong at the date / the reverse; reverted to baseline: base$^{\dagger}$ in Table~\ref{tab:results}.}
\label{tab:pvals}
\end{table*}

\section{Churn between consecutive days}\label{app:adjacent}
\begin{table}[!htbp]\centering\small\setlength{\tabcolsep}{4pt}
\begin{tabular}{@{}lrrr@{}}
\toprule
Model & One day & Plausible & Batch \\
\midrule
Llama-3.2-3B   & 4.2  & 8.0  & 5.5 \\
Llama-3.1-8B   & 3.6  & 6.3  & 4.7 \\
SmolLM3-3B     & 16.0 & 15.9 & 16.4 \\
Granite-3.3-8B & 6.1  & 5.9  & 4.9 \\
Mistral-24B    & 2.7  & 2.7  & 2.9 \\
gpt-oss-20B    & 8.9  & 9.5  & 8.6 \\
\bottomrule
\end{tabular}
\caption{\textbf{Correctness churn between consecutive days} (\% of 96 TEST items, mean over each model's four
cells). One day: the three consecutive-day pairs; Plausible: the 36 pairs of nine plausible dates (\S\ref{sec:cost});
Batch: the baseline date at batch size 8 vs.\ 32.}
\label{tab:adjacent}
\end{table}
To measure what a re-run on the next day changes, we score, per cell, three pairs of consecutive dates in the model's
own format: the baseline and the day after it, \texttt{10}/\texttt{11 Feb 2025}, and \texttt{25}/\texttt{26 Nov 2025}.
The earlier date of each pair is one of the nine plausible dates of \S\ref{sec:cost}, and within a pair the prompts
differ in exactly one token. Generation and scoring follow the enumeration (Appendix~\ref{app:repro}); the baseline
regenerates byte-for-byte in all 24 cells. Correctness churn between consecutive days (mean of the three pairs) has a
median of $4.3\%$ over cells ($95\%$ CI $3.5$--$5.9$; range $1.0$--$22.9\%$), against $6.6\%$ for the 36
plausible-date pairs and $4.2\%$ for the batch-size control (Table~\ref{tab:adjacent}). It is lower than the
plausible-pair churn in 17 of 24 cells (6 higher, 1 tied; Wilcoxon signed-rank $p{<}0.01$), and we detect no difference from
the batch-size control ($p{\approx}0.5$). The gap is the Llama models': a one-day change flips about half as many of their
items as two plausible dates (fewer in all eight Llama cells), consistent with their output depending largely on the
month (Appendix~\ref{app:inwindow}); for the other four models it flips as many. Between consecutive days the final
answer is not byte-identical on a median $62\%$ of items ($75\%$ between plausible dates).

\section{In-window worst case}\label{app:inwindow}
The date grammar admits $1116$ day/month/year combinations in 2024--2026; we drop the $20$ impossible ones (e.g.,
30 February) and rank the remaining $1096$ real dates of each cell by the decay-weighted output KL from the baseline on
the 32 TRAIN prompts (forward passes only). The six highest are scored on the 96 TEST items with exact McNemar tests
against the baseline (Table~\ref{tab:inwindow}); the baseline, re-scored in the same run, reproduces the main run
byte-for-byte in all 24 cells. Its KL in the screen, from a batched forward pass against its own batch-1 pass, is the bf16 noise floor, against which the screen's top date stands
out by $23$--$60\times$ for Llama-3.2-3B, $7.5$--$39\times$ for Llama-3.1-8B, and $1.8$--$14\times$ for Granite,
Mistral, and gpt-oss, but only $1.1$--$1.3\times$ for SmolLM3, whose shortlist is therefore effectively random. Had the 20 impossible combinations been kept, they would take
Granite's top six to eight KL ranks in all four cells, as its impossible optimized dates in Table~\ref{tab:results} suggest. No
shortlisted date survives Bonferroni correction over the 144 comparisons (six dates in each of 24 cells), nor does any loss at the eight random dates survive
Bonferroni over their 192 comparisons (smallest unadjusted $p{=}0.035$, SmolLM3$\cdot$MMLU at
\texttt{10 January 2024}, $9.4$\,pp). Even unadjusted, no shortlisted date and only 2 of the 192 random-date comparisons reach
$p{<}0.05$, one of them a gain. The largest shortlisted loss, gpt-oss$\cdot$IFEval at \texttt{2024-02-02}
($12.5$\,pp, $p{=}0.050$), is the non-termination failure of \S\ref{sec:infer}: generations without a final answer
rise from $28$ to $42$ of 96. Table~\ref{tab:results}'s plausible block pools these six dates with the baseline and
the eight random 2024--2026 dates (14 distinct dates for SmolLM3$\cdot$IFEval, whose sixth shortlisted date,
\texttt{17 June 2024}, is also one of its random dates). Outside three cells the shortlist is no more damaging than random
draws: in the other 21 the mean loss per date is $-0.5$\,pp at the random dates and $-0.7$\,pp at the shortlisted ones
(higher for the shortlist in 8), and over all 24 cells the larger worst loss is a random date's in 11 and a
shortlisted date's in 7 (6 ties). The exceptions are Llama-3.2-3B's MMLU and IFEval and gpt-oss$\cdot$IFEval, where the
shortlisted dates cost $2.8$--$5.6$\,pp on average against $-1.2$ to $2.1$\,pp for the random dates.

\begin{figure*}[t]
\centering
\includegraphics[width=0.92\textwidth]{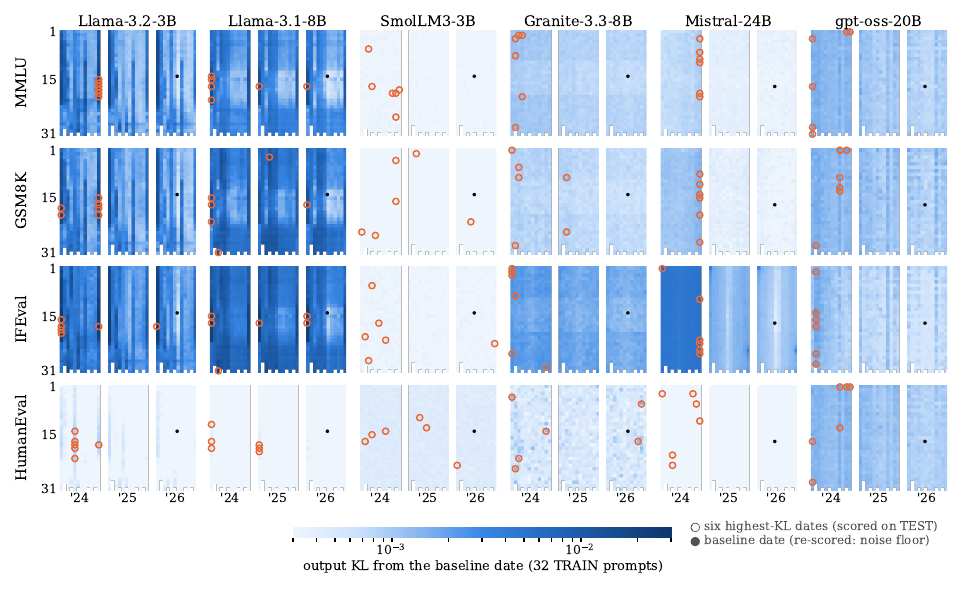}
\caption{\textbf{Which in-window dates shift each model's output.} One calendar per cell (columns: the 36
months of 2024--2026; rows: day of the month); each square is a real date shaded by its output KL from the baseline
(log scale shared across panels, clipped to the range $3{\times}10^{-4}$ to $3{\times}10^{-2}$; blank: impossible dates). Rings: the six highest-KL dates
(Table~\ref{tab:inwindow}); dot: the baseline date, at the noise floor.}
\label{fig:klcal}
\end{figure*}

\begin{figure*}[t]
\centering
\includegraphics[width=\textwidth]{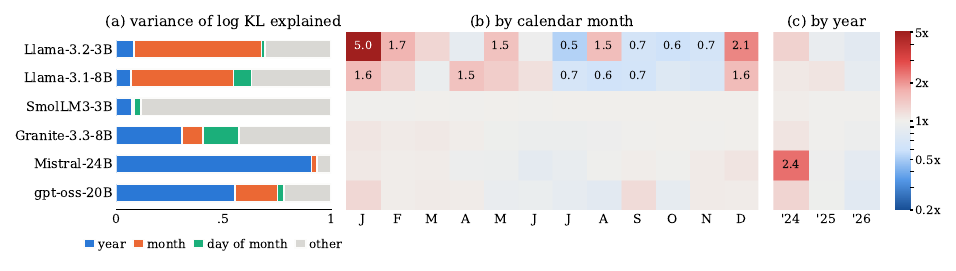}
\caption{\textbf{The model decides which part of the date matters.} Over the $1096$ real in-window dates
per cell: \textbf{(a)}~share of the variance of log output KL explained by year, month, and day (mean over
benchmarks); \textbf{(b,\,c)}~median KL per month and per year relative to the cell's median date (geometric mean over benchmarks; labeled where the
ratio is at least $1.45\times$ or at most $1/1.45$).}
\label{fig:datepart}
\end{figure*}

\paragraph{Which part of the date matters.} Figure~\ref{fig:klcal} maps the screen and Figure~\ref{fig:datepart}
summarizes it. Which part of the date moves the output depends on the model. For the two Llama models the month
explains $41$--$63\%$ of the variance of log KL (the year $3$--$11\%$): Llama-3.2-3B's January dates shift its output
$5.0\times$ as much as its median date and its December dates $2.1\times$, while July, the baseline's month, is lowest
($0.5\times$). For Mistral the year explains $83$--$95\%$ (2024: $2.4\times$), for gpt-oss mostly the year as well
($44$--$67\%$), and for SmolLM3 nothing does, its KL being at the noise floor. The shortlists inherit this
structure: 40 of the Llama models' 48 shortlisted dates fall in mid-December or mid-January, and all of Mistral's and
gpt-oss's fall in 2024 (19 of Mistral's 24 in November--December). Llama's peak months adjoin the December 2023 cutoff
its template states (Appendix~\ref{sec:cutoff}), but January is also the month farthest from the July baseline, and
we cannot separate the two explanations without a second baseline.

\begin{table*}[!htbp]\centering\small\setlength{\tabcolsep}{4pt}
\begin{tabular}{@{}ll r lr rrr r@{}}
\toprule
 & & & \multicolumn{5}{c}{Worst of the six highest-KL dates} & \\
\cmidrule(lr){4-8}
Model & Bench & KL$_{\max}$/floor & date & rank & $\Delta$ & R$\to$W/W$\to$R & $p$ & min $p$ \\
\midrule
Llama-3.2-3B & MMLU & 55.4 & \texttt{15 Dec 2024} & 4 & +6.2 & 8/2 & 0.109 & 0.109 \\
 & GSM8K & 53.2 & \texttt{17 Dec 2024} & 5 & +1.0 & 6/5 & 1.000 & 0.754 \\
 & IFEval & 59.8 & \texttt{20 Jan 2024} & 2 & +4.2 & 8/4 & 0.388 & 0.388 \\
 & HumanEval & 23.2 & \texttt{18 Dec 2024} & 3 & $-$1.0 & 3/4 & 1.000 & 0.289 \\
\addlinespace[2pt]
Llama-3.1-8B & MMLU & 38.0 & \texttt{14 Jan 2024} & 5 & +2.1 & 6/4 & 0.754 & 0.688 \\
 & GSM8K & 27.9 & \texttt{17 Jan 2024} & 1 & +1.0 & 3/2 & 1.000 & 0.125 \\
 & IFEval & 39.4 & \texttt{17 Jan 2026} & 1 & +0.0 & 7/7 & 1.000 & 0.629 \\
 & HumanEval & 7.5 & \texttt{19 Jan 2024} & 1 & +1.0 & 2/1 & 1.000 & 1.000 \\
\addlinespace[2pt]
SmolLM3-3B & MMLU & 1.2 & \texttt{19 October 2024} & 3 & +6.2 & 11/5 & 0.210 & 0.210 \\
 & GSM8K & 1.2 & \texttt{16 November 2024} & 2 & +3.1 & 7/4 & 0.549 & 0.549 \\
 & IFEval & 1.3 & \texttt{28 March 2024} & 5 & +6.2 & 11/5 & 0.210 & 0.210 \\
 & HumanEval & 1.1 & \texttt{14 August 2024} & 2 & +4.2 & 12/8 & 0.503 & 0.503 \\
\addlinespace[2pt]
Granite-3.3-8B & MMLU & 2.1 & \texttt{April 20, 2024} & 5 & +1.0 & 2/1 & 1.000 & 0.344 \\
 & GSM8K & 1.8 & \texttt{January 01, 2024} & 1 & +1.0 & 4/3 & 1.000 & 1.000 \\
 & IFEval & 3.2 & \texttt{November 30, 2024} & 2 & $-$1.0 & 4/5 & 1.000 & 0.092 \\
 & HumanEval & 2.6 & \texttt{November 06, 2026} & 1 & +0.0 & 1/1 & 1.000 & 1.000 \\
\addlinespace[2pt]
Mistral-24B & MMLU & 3.3 & \texttt{2024-12-09} & 2 & +2.1 & 3/1 & 0.625 & 0.500 \\
 & GSM8K & 5.9 & \texttt{2024-12-20} & 6 & +0.0 & 2/2 & 1.000 & 0.500 \\
 & IFEval & 14.0 & \texttt{2024-12-26} & 3 & +3.1 & 7/4 & 0.549 & 0.549 \\
 & HumanEval & 4.4 & \texttt{2024-11-06} & 1 & $-$2.1 & 0/2 & 0.500 & 0.500 \\
\addlinespace[2pt]
gpt-oss-20B & MMLU & 2.9 & \texttt{2024-01-03} & 6 & +1.0 & 3/2 & 1.000 & 0.375 \\
 & GSM8K & 2.4 & \texttt{2024-09-01} & 3 & +0.0 & 1/1 & 1.000 & 0.500 \\
 & IFEval & 4.7 & \texttt{2024-02-02} & 4 & +12.5 & 22/10 & 0.050 & 0.050 \\
 & HumanEval & 2.5 & \texttt{2024-11-01} & 6 & +2.1 & 3/1 & 0.625 & 0.625 \\
\bottomrule
\end{tabular}

\caption{\textbf{The six highest-KL in-window dates per cell, scored on TEST} ($n{=}96$). KL$_{\max}$/floor: the
top date's TRAIN KL over the re-scored baseline's. Worst: the shortlisted date with the largest loss (rank: its KL
rank among the $1096$; ties broken by smaller $p$, then by higher KL). $\Delta$: baseline${-}$date accuracy (pp, positive${=}$loss);
$p$: exact McNemar. min $p$: smallest $p$ among the six, either direction.}
\label{tab:inwindow}
\end{table*}

\section{Jailbreak study details}\label{app:jb}
\paragraph{Prompts and reference date.} The 786 prompts are all prompts in a stored jailbreak set for Llama-3.2-3B
whose response, generated under the pinned date \texttt{26 Apr 2026}, scored above $0.6$ under the StrongREJECT
fine-tuned evaluator~\citep{souly2024strongreject}, a margin over the $0.5$ success threshold below (the 206 stored
successes scoring $0.5$--$0.6$ are excluded). They were produced with EasyJailbreak~\citep{zhou2024easyjailbreak}
and GCG~\citep{gcg} on AdvBench behaviors~\citep{gcg}: ReNeLLM~\citep{ding2024renellm} (375),
GCG (147), GPTFuzzer~\citep{yu2023gptfuzzer,yu2024llmfuzzer} (122), DeepInception~\citep{li2023deepinception} (74),
CodeChameleon~\citep{lv2024codechameleon} (64), and AutoDAN~\citep{liu2024autodan} (4). Table~\ref{tab:jb} gives survival by family.
\paragraph{Protocol.} Each prompt is a single user turn with no custom system prompt, rendered with the model's
chat template and an explicit \texttt{date\_string}; greedy decoding, up to 4096 new tokens, batch size 1, bf16, in
the software environment that produced the stored responses (Transformers 4.48.2, PyTorch 2.7.1); an upgrade to
Transformers 4.55.0 alone changes 10 of 32 responses (and 2 labels) in a pilot, and batch size 4 (left-padded) changes 28
and flips 9 of 32 stored successes, so both are held fixed. A response succeeds when the StrongREJECT fine-tuned score is $\ge0.5$ (judge
batch size 1). Dates: the baseline \texttt{14 Jul 2026}, eight random 2024--2026 dates (a separate draw from the enumeration's eight of Appendix~\ref{app:repro}), four random dates outside
that window, Llama-3.2-3B's VAL-selected IFEval and MMLU dates (\texttt{28 Sep 1963}, \texttt{25 Feb 1941}), and the clock dates
\texttt{01 Jan 1970} and \texttt{01 Jan 2000}; all render to the same
token count. For Llama-3.1-8B, survival is measured relative to its own regeneration of the same prompts at
\texttt{26 Apr 2026}, among the 472 that succeed there.

\begin{table}[!htbp]\centering\small\setlength{\tabcolsep}{3pt}
\begin{tabular}{@{}lrccc@{}}
\toprule
Family & $n$ & Plausible & 1963 & 1970 \\
\midrule
ReNeLLM        & 375 & 81--93 & 54 & 62 \\
GCG            & 147 & 65--76 & 12 & 24 \\
GPTFuzzer      & 122 & 43--92 & 15 & 21 \\
DeepInception  &  74 & 58--93 &  1 & 23 \\
CodeChameleon  &  64 & 52--72 & 48 & 64 \\
AutoDAN        &   4 & 0--75  &  0 &  0 \\
\midrule
All            & 786 & 68--85 & 34 & 45 \\
\bottomrule
\end{tabular}
\caption{\textbf{Jailbreak survival on Llama-3.2-3B by attack family} (\% of stored successes that still succeed).
\emph{Plausible}: range over the baseline and eight random 2024--2026 dates. 1963, 1970: \texttt{28 Sep 1963} (VAL-selected IFEval) and \texttt{01 Jan 1970}.}
\label{tab:jb}
\end{table}

\section{XSTest study details}\label{app:xstest}
\paragraph{Prompts and protocol.} We use the XSTest prompt file \citep{rottger2024xstest} from its official repository
(\texttt{xstest\_prompts.csv} at commit \texttt{d7bb5bd}): 100 of its 250 safe prompts (10 per type) and 100 of its 200
unsafe contrast prompts (12--13 for each of its 8 types), drawn with seed 20261006. Each prompt is a single user turn with no custom system prompt, so Granite's and Mistral's default system
messages, which carry the date, are rendered; we verified the date in every rendered prompt. Generation follows each
model's main-study settings (greedy, bf16, batch 32 with the same batch composition at every date, up to 4096 new
tokens). Dates: each model's baseline (Table~\ref{tab:grammar}) and the jailbreak study's eight random
in-window, four random out-of-window, and four named dates, each in the model's own format (16 dates for Mistral and
gpt-oss, whose baseline is one of the in-window dates). The baseline repeated in the same process reproduces
byte-for-byte for all six models (in a smoke run on the first 64 prompts); a noise-floor control re-runs the baseline date in a reshuffled batch order.
\paragraph{Classification.} Only the final answer is classified; a generation with no final answer is counted
separately (gpt-oss: $6$--$18\%$ of generations, SmolLM3: $0.5$--$4\%$, varying with the date). The primary label is Qwen3Guard-Gen-8B's
refusal output \citep{zhao2025qwen3guard}. In our validation against the 2,250 human-labeled responses released with
XSTest (full or partial refusal counted as refusal), it reaches accuracy $0.918$ and $\kappa{=}0.84$; XSTest's string
match, our secondary label, reaches $0.884$ and $\kappa{=}0.77$; we apply it after normalizing typographic apostrophes, without which
it misses every gpt-oss refusal. Refusal rates are over all 100 prompts per label, a generation with no final answer counting as
no refusal. Tests are exact McNemar against the baseline on the prompts with a final answer at both dates, Holm-corrected over
each model's dates (not across models); coding refusal against every other outcome on all 100 prompts gives the same decisions. With 100
prompts per label, failure to reject does not rule out small effects. Table~\ref{tab:xstest} gives the per-model rates. On the 100 unsafe prompts, no date changes any model's refusal
rate statistically significantly (Holm, primary label): no model's rate falls more than 6\,pp below its baseline
($62$--$99\%$ across models; the two labels disagree on Granite's and SmolLM3's unsafe-prompt answers, $\kappa{\le}0.3$ on our generations, so those levels depend on the label), and the only changes with unadjusted $p{<}0.05$ are rises (Llama-3.2-3B: $88\%$ at baseline, $94$--$96\%$ at seven of its eight
out-of-window and named dates). Under the string-match label, Mistral's unsafe-prompt refusal falls at 1963 and 1941
($83\%$ to $73\%$, Holm $p{=}0.03$), which the primary label does not show.

\begin{table*}[!htbp]\centering\small\setlength{\tabcolsep}{3pt}
\begin{tabular}{@{}lrrrrrrr@{}}
\toprule
Model & base & plausible & out & 1963 & 1941 & kept & churn / max / noise \\
\midrule
Llama-3.2-3B & 6 & 6--9 & 9--12 & 18$^{\ast}$ & 22$^{\ast}$ & 96.8 & 2.5 / 4.5 / 0.5 \\
Llama-3.1-8B & 9 & 9--10 & 9--28$^{\ast}$ & 31$^{\ast}$ & 24$^{\ast}$ & 97.8 & 1.2 / 2.5 / 0.0 \\
SmolLM3-3B & 2 & 0--3 & 0 & 0 & 2 & 81.7 & 10.2 / 14.0 / 8.0 \\
Granite-3.3-8B & 18 & 17--20 & 17--21 & 19 & 17 & 97.6 & 1.3 / 3.0 / 0.5 \\
Mistral-24B & 12 & 10--12 & 10--12 & 22 & 24$^{\ast}$ & 100.0 & 1.0 / 2.0 / 1.0 \\
gpt-oss-20B & 10 & 8--12 & 6--12 & 12 & 10 & 49.3 & 14.0 / 17.0 / 9.0 \\
\bottomrule
\end{tabular}
\caption{\textbf{XSTest over-refusal by date} (\% of 100 safe prompts refused). base: the baseline date; plausible: the baseline and
random 2024--2026 dates; out: four random out-of-window dates; kept: share of the safe prompts answered at baseline
that are answered at every plausible date (a refusal or no final answer counts as not answered); churn / max / noise: mean and max
share of the 200 prompts whose outcome (refusal, other answer, or no final answer) differs between two plausible dates,
and between the baseline and its reshuffled re-run; for gpt-oss most of this churn is in whether a final answer appears
(on prompts answered at both dates its mean is $1.5\%$, and SmolLM3's $7.1\%$). $^{\ast}$Holm $p{<}0.05$ (in a range: for at least one date). 1963, 1941: \texttt{28 Sep 1963} and \texttt{25 Feb 1941} in each model's format.}
\label{tab:xstest}
\end{table*}

\section{Activation-monitor study details}\label{app:probes}
\paragraph{Probes.} For each of Llama-3.2-3B-Instruct's 28 layers, a logistic-regression and an MLP probe read the
residual stream at the last prompt token and flag harmful and jailbreak prompts against benign ones. The probe data
are 6,482 prompts (Dolly-derived benign prompts, \citealp{conover2023dolly}; AdvBench, HarmBench, and XSTest prompts; and AdvBench jailbreak
variants, excluding AutoDAN and cipher attacks), split 2,649/1,277/2,556 into training, validation, and test sets
grouped by base query (seed 42), so no base query appears in two splits. CodeChameleon and ReNeLLM record their
transformed request as the base query, so the AdvBench request behind 186 of their test prompts appears, untransformed, in training. Features are standardized; the logistic probe is linear and the MLP has one 64-unit
ReLU layer, both trained for one epoch of Adam (learning rate $10^{-3}$, weight decay $10^{-5}$, batch 64) with a
class-balanced loss. Training prompts were rendered
at \texttt{26 Apr 2026}, and each probe is thresholded at $1\%$ false positives on validation prompts rendered at
that date. We re-render the held-out test prompts (1,202 benign, 346 harmful, 1,008 jailbreak) under 25 other dates:
the jailbreak study's 17, and the eight in-window dates with the highest geometric-mean output KL over
Llama-3.2-3B's four benchmark cells (measured against \texttt{14 Jul 2026}, the screen's baseline). Every date renders to the same token
count and positions, and the training date reproduces the stored features and predictions bit-for-bit.
\paragraph{Analysis.} Averaged over layers 1--14, benign false positives stay at $0.6$--$1.3\%$ at every date, so we
report the deeper half (layers 15--28), where the large shifts are. This grouping is exploratory, chosen after
inspecting the layerwise results; the Holm correction below covers the dates, not this choice. Shallow layers are not inert: logistic regression's
validation-selected layer 13 rises from $0.6\%$ to $2.4$--$2.7\%$ at 1941 and 1963, and at the eight high-KL dates
the shallow mean falls about as much as the deep one. Tests are paired Wilcoxon tests against the training date, Holm-corrected over the
25 dates: on each benign prompt's share of deep layers that flag it (Table~\ref{tab:probes}), and on each harmful
prompt's share that misses it, whose rise at the eight high-KL dates is significant for both probe types (Holm
$p{\le}0.026$). Across the 17 in-window dates (the baseline, eight random plausible, and eight high-KL dates), output KL
and the shift in benign false positives correlate at Spearman $\rho{=}{-}0.76$ (logistic regression) and ${-}0.70$
(MLP), but within the baseline and eight random dates $\rho{=}0.15$: the correlation reflects the high-KL group as a
whole, whose shifts all lie below those of these nine dates and whose eight dates cluster in two calendar spots, so they are not
eight independent observations.

\begin{table*}[!htbp]\centering\small\setlength{\tabcolsep}{2.5pt}
\begin{tabular}{@{}lll@{}}
\toprule
Dates & FPR (worst layer) & harmful miss \\
\midrule
training (26 Apr 2026) & 1.0 (1.7) / 0.8 (1.7) & 2.4 / 1.5 \\
baseline + 8 random plausible & 0.8--1.3 (${\le}1.9$) / 0.7--1.0 (${\le}2.3$) & 1.4--2.6 / 1.3--1.7 \\
8 highest-KL plausible$^{\ast}$ & 0.55--0.60 (${\le}1.2$) / 0.49--0.53 (${\le}1.3$) & 3.7--4.4 / 2.7--3.1 \\
4 random out-of-window$^{\ast}$ & 1.6--2.8 (${\le}5.2$) / 1.2--2.2 (${\le}7.7$) & 1.0--1.2 / 1.0--1.1 \\
01 Jan 1970 / 2000$^{\ast}$ & 1.9--2.3 (${\le}4.7$) / 1.4--1.8 (${\le}6.1$) & 1.2--1.3 / 1.1 \\
25 Feb 1941$^{\ast}$ & 9.6 (29.2) / 8.3 (27.3) & 0.6 / 0.6 \\
28 Sep 1963$^{\ast}$ & 11.6 (35.0) / 10.4 (34.7) & 0.5 / 0.4 \\
\bottomrule
\end{tabular}
\caption{\textbf{Deep-layer (15--28) probe rates by date} (\%; logistic regression / MLP). FPR: benign prompts flagged,
mean over layers (worst layer in parentheses); miss: harmful prompts not flagged, mean over layers. Jailbreak misses
are at most $0.3\%$ at every date and deep layer. $^{\ast}$All dates in the row Holm-significant (benign FPR). In the baseline + 8 random row, benign FPR is also
Holm-significant at 23 Feb 2026 (both probes, up $0.2$--$0.3$\,pp) and 20 Apr 2025 (logistic regression, down $0.1$\,pp).}
\label{tab:probes}
\end{table*}

\end{document}